%% file: main.tex
\documentclass[10pt,twocolumn,letterpaper]{article}

\newif\ifarxivversion
\arxivversiontrue
\ifarxivversion
  \usepackage[pagenumbers]{cvpr}
  \makeatletter
  \providecommand{\@LN@col}[1]{}
  \patchcmd{\@maketitle}{\vskip .375in}{\vskip 0pt}{}{}
  \patchcmd{\@maketitle}{\vspace*{24pt}}{\vspace*{10pt}}{}{}
  \patchcmd{\@maketitle}{\vspace*{12pt}}{\vspace*{6pt}}{}{}
  \makeatother
\else
  \usepackage[review]{cvpr}
\fi
\input{preamble}

\usepackage{chapterbib}
\definecolor{cvprblue}{rgb}{0.21,0.49,0.74}
\usepackage[pagebackref,breaklinks,colorlinks,allcolors=cvprblue]{hyperref}
\def\confName{CVPR}
\def\confYear{2027}
\title{OmniScore-Bench: Evaluating Vision-Language Models\\on Musical Score Reading}
\ifarxivversion
  \usepackage{marvosym}
  \author{%
    Congren Dai\textsuperscript{1,2},
    Zhancheng Guo\textsuperscript{1,2},
    Danni Zhao\textsuperscript{1},\\
    Wenchu Wang\textsuperscript{1,2},
    Xiaobing Li\textsuperscript{1},
    and Maosong Sun\textsuperscript{1,2\Letter}\\[5pt]
    \textsuperscript{1}Central Conservatory of Music\quad
    \textsuperscript{2}Tsinghua University\\
    {\tt\small congren.dai@mail.ccom.edu.cn\quad sms@tsinghua.edu.cn}%
  }
  \hypersetup{pdfauthor={Congren Dai, Zhancheng Guo, Danni Zhao, Wenchu Wang, Xiaobing Li, Maosong Sun}}
  \AddToHook{cmd/maketitle/after}{%
    \begingroup
    \footnotetext[1]{Corresponding author.}%
    \endgroup
  }
\else
  \author{Anonymous submission}
  \hypersetup{pdfauthor={Anonymous submission}}
\fi
\begin{document}
\include{sec/paper}
\include{supplementary}
\end{document}

%% file: preamble.tex
\usepackage{microtype}
\usepackage{mathtools}
\usepackage{tabularx}
\usepackage{array}
\usepackage{multirow}
\usepackage{pifont}
\newcommand{\evalyes}{\textcolor{black}{\ding{51}}}
\newcommand{\evalno}{\textcolor{black}{\ding{55}}}
\usepackage{tikz}
\usetikzlibrary{arrows.meta,positioning,calc}
\newcommand{\bench}{OmniScore-Bench\xspace}
\definecolor{omrblue}{HTML}{2B536E}
\definecolor{omrorange}{HTML}{A76935}
\newcolumntype{Y}{>{\raggedright\arraybackslash}X}

%% file: sec/paper.tex
\maketitle
\input{sec/0_abstract}

\begin{figure*}[t]
\centering
\includegraphics[width=\linewidth]{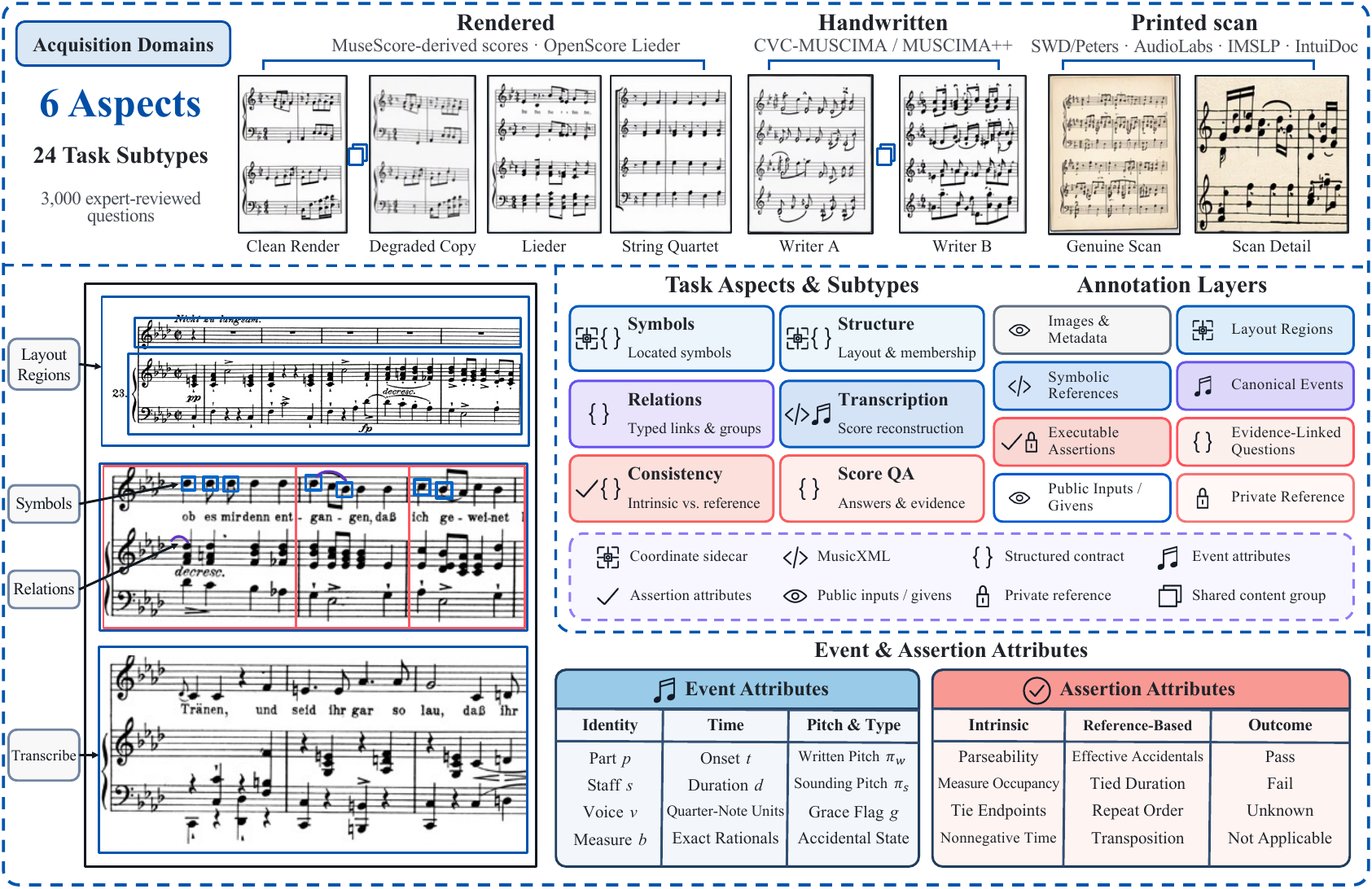}
\caption{\textbf{Six aspects of musical score reading for vision-language models.} The evaluation comprises 3,000 questions, with 1,000 each from Rendered, Printed scan and Handwritten inputs and 125 per subtype. The tasks connect visible symbols to structure, relations, transcription, musical consistency and evidence-linked answers. Event and assertion attributes describe reference annotations.}
\label{fig:benchmark}
\end{figure*}
\input{sec/1_intro}

\section{Related Work}
\label{sec:related}
\paragraph{Optical music recognition.}
OMR encompasses image analysis, notation reconstruction and interpretation~\cite{calvo2021understanding,byrd2015testbed}. DeepScores and DeepScoresV2 provide dense object annotations~\cite{tuggener2018deepscores,tuggener2021deepscoresv2}; CVC-MUSCIMA supplies handwritten images, while MUSCIMA++ adds symbols and notation relationships~\cite{fornes2012cvc,hajic2017muscima}. Monophonic recognition datasets and pianoform datasets support transcription from single staves to multiple staves and full pages~\cite{calvo2018primus,rios2023pianoform,rios2026fullpage}. These resources cover complementary visual and symbolic levels, from individual objects to full scores. Our questions expose several of these annotation levels through a common multimodal interface. Symbol positions reveal localisation errors; staff assignments and relation annotations additionally expose whether a system connects the musical objects correctly.

\paragraph{Score comparison.}
SMB standardises OMR evaluation on scanned pages and introduces OMR-NED for detailed notation comparison~\cite{martinez2025smb}. Foscarin et al. compare score files~\cite{foscarin2019diff}, while Muse OMR Benchmark supplies image--encoding pairs and evaluation tools~\cite{musegroup2025benchmark}. We build on established transcription measurements rather than propose a replacement distance. OmniScore-Bench adds separately scored questions about image regions, object membership, relations and musical interpretation. This distinction matters because transcription correctness, localisation accuracy and successful evidence citation require different predictions. Table~\ref{tab:positioning} compares the evaluation focus of established OMR resources with our six aspects.

\paragraph{Document understanding and evaluation.}
Document benchmarks also examine capabilities beyond transcription. OCRBench v2 evaluates visual text localisation, extraction and reasoning~\cite{fu2025ocrbenchv2}, while OmniDocBench analyses document parsing across sources, tasks and attributes~\cite{ouyang2025omnidocbench}. The olmOCR system introduces a companion benchmark with unit tests for content and structure~\cite{poznanski2025olmocr}; olmOCR~2 uses such tests as training rewards~\cite{poznanski2025olmocr2}. These works motivate examining specific document-reading operations. Musical notation additionally requires pitch, duration, simultaneity, ties and repeat interpretation. OmniScore-Bench studies these musical capabilities alongside localisation and transcription, with separate task-level measurements.

\begin{figure}[t]
\centering
\includegraphics[width=\linewidth]{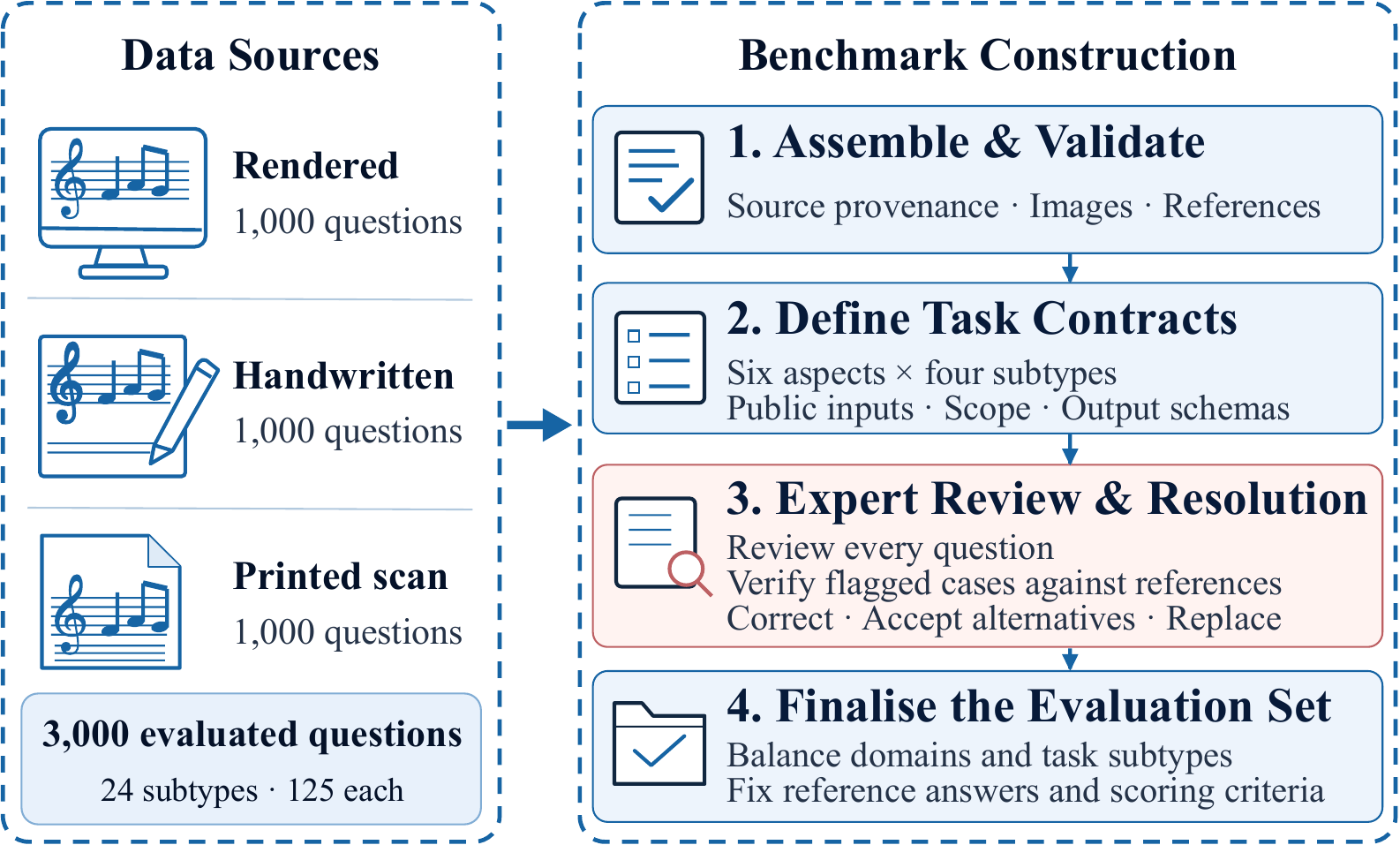}
\caption{\textbf{References are bound to visible task scope.} Sources and annotations define candidate questions; automatic checks and source-based resolution of flagged cases precede the final evaluation set. The evaluation uses one question per request.}
\label{fig:construction}
\end{figure}

\begin{table*}[t]
\centering\small
\caption{\textbf{Evaluation focus across score-reading datasets and benchmarks.} \evalyes\ indicates explicit support for at least one target in an aspect; \evalno\ indicates it is not a dedicated evaluation focus. A tick does not imply coverage of all OmniScore subtypes. T4 includes musical-content extraction in MUSCIMA++; Muse's structure and relation checks operate on symbolic scores. T5 denotes separate musical-consistency queries; T6 covers score QA, with explicit evidence scoring in OmniScore.}
\label{tab:positioning}
\input{tables/omr_focus_matrix}
\end{table*}

\section{Benchmark Construction}
\label{sec:benchmark}
\subsection{Sources, scope and sampling}
The evaluation contains 1,000 questions each from Rendered, Printed scan and Handwritten inputs, totalling 3,000. Each of the six aspects contains four subtypes and each subtype contains 125 questions. 

Rendered inputs come from MuseScore-related encodings and OpenScore Lieder~\cite{musegroup2025benchmark,gotham2022lieder}. Handwriting comes from CVC-MUSCIMA and MUSCIMA++~\cite{fornes2012cvc,hajic2017muscima}. Printed scans combine the Schubert Winterreise material, measure-region annotations, additional score scans and an accidental-detection collection~\cite{weiss2021winterreise,weiss2024winterreise,zalkow2019measures,intuidoc2018accidentals}. The Supplementary Material (SM) gives collection counts in Table~\ref{tab:sources}.

An item declares the image regions and musical extent that its answer must cover. Multipage transcription reconstructs the supplied pages in order, with the requested extent defined by the visible input. A question can include target identifiers, structural assignments or an interpretation rule that defines the requested operation. Reference annotations specify the answer within that scope. For example, supplying the relevant measure isolates a pitch-selection question from the additional problem of locating that measure on a page.

\subsection{Reference construction and review}
Task construction starts from image--annotation correspondence (Fig.~\ref{fig:construction}). Symbol and layout annotations support geometric questions; symbolic encodings provide pitch, duration and structural references where available. We instantiate questions only for supported annotation layers and specify the relevant visible scope. Automatic checks test whether the question has a well-defined answer and whether its image and reference cover the same musical scope. For printed scans, correspondence between the visible notation and symbolic encoding is part of reference verification.

Two domain experts, each with over 20 years of experience, independently review all 3,000 questions while blind to model responses and each other's judgements, with an initial agreement of 92.50\%; a third equally experienced expert reviews disagreements and determines the final reference answers.

All models receive the same 3,000 evaluation questions. The study evaluates pretrained models without task-specific training or a development/test split.

\subsection{Six aspects of score reading}
\paragraph{T1: Symbols.}
The model returns symbol classes and normalised image boxes. The subtypes cover noteheads and rests, clefs and accidentals, rhythmic marks such as stems and beams, and expressive or articulatory symbols. Class-aware one-to-one matching at an intersection-over-union threshold of 0.5 defines the strict localisation score. A separate one-pixel tolerance analysis measures sensitivity to coordinate precision.

\paragraph{T2: Structure.}
Questions concern page and system organisation, measure regions, part/staff/voice membership and structural continuation. Geometric outputs are scored as regions, while membership and continuation use their declared associations. Some questions supply object identities or structural givens, so success measures the requested association rather than independent recovery of every upstream component.

\paragraph{T3: Relations.}
The four subtypes examine ties and slurs, note grouping, temporal relationships and additional typed connections. A tie and a slur are different relations even when their curves look similar. Slur matching permits specified chord-member equivalences, while tie matching remains pitch-sensitive. Exact answers and relation or grouping measurements are reported separately.

\paragraph{T4: Transcription.}
All 500 transcription questions require MusicXML. T4.1 and T4.2 cover bounded excerpts, T4.3 covers the displayed single-page scope, and T4.4 covers an ordered multipage input. Explicit inherited context or grouping instructions are retained in the prompt when required. Some transcription metrics compare ABC representations derived from the submitted MusicXML.

\paragraph{T5: Musical consistency.}
Questions target timing and metrical interpretation, written and sounding pitch, sounding events and bounded repeat traversal. Each question specifies the relevant events, transformation assumptions and treatment of ambiguity. These tasks test the musical consequences of notation, including the duration implied by valid ties, independently of full-score transcription.

\paragraph{T6: Score question answering.}
The subtypes ask about visible facts, local counting and retrieval, conditional musical analysis and fragment transformations. Counting and retrieval retain distinct output and scoring contracts. Answers include required evidence, either image regions or supplied region identifiers. Joint success checks the answer and its applicable evidence requirement. The task-specific givens distinguish these measurements from unconstrained music reasoning.

\section{Experiments}
\label{sec:experiments}
\subsection{Experimental Setup}
\label{sec:protocol}
\subsubsection{Models, inference and answer handling}
We evaluate twelve vision-language models (VLMs) from seven developer families: DeepSeek V4.1 Flash~\cite{deepseek2026flash1017}; Qwen 3.7 Plus and 3.8 Max~\cite{qwen2026plus37,qwen2026max38}; Kimi K3 and K3.1 Pro~\cite{moonshot2026k3,moonshot2026k31pro}; GLM 5.3 Flash~\cite{zai2026glm53flash}; Claude Fable 5.1 and Opus 5.5~\cite{anthropic2026fable51,anthropic2026opus55}; Gemini 3.8 Flash and 4 Argon~\cite{google2026gemini38flash,google2026gemini4argon}; and GPT-6.1 Sol and GPT-6 Astra~\cite{openai2026sol61,openai2026astra}. Each request contains one question and its image input, with the same task instructions across models. Each model answers every question independently, without access to other models' answers or the reference annotations. The inputs specify the requested scope and answer format. 

The strict evaluator applies the required JSON or MusicXML contract. All MusicXML answers undergo the same semantics-preserving representation normalization before validation and metric computation. Model-generated truncation, refusal and invalid or empty answers count as output failures. Communication failures may be retried. A supplemental pass removes only one permitted outermost Markdown code fence from an otherwise complete eligible answer, then repeats the original validation and scoring. It does not correct musical content or extract a preferred fragment from explanatory text. No LLM judge is used. SM Sec.~\ref{sec:question-prompts} provides the question prompts.

\subsubsection{Comparison design}
The main comparison contains 36,000 model--question pairs, with 1,905 distinct original images and 4,535 original-image occurrences. The questions form 616 source-content groups, defined by source identity. Non-transcription scores average within and then across these groups.

The one-pixel and outer-fence analyses reuse the main comparison's answers and alter only the specified scoring condition, without changing examples or references. 

\subsubsection{Metrics and aggregation}
Localisation uses class-aware detection F1. Structural and relational tasks use the matching, membership or exact-answer rules specified for their outputs. Musical questions compare the requested pitch, time, sounding-event or traversal fields. Question answering reports answers and evidence separately as well as joint success. Each subtype retains its own measurements; we do not average F1, exact match and edit distances into an overall leaderboard.

For a metric with eligible item set $E_t$ in subtype $t$, the primary non-transcription summary averages within source-content groups and then equally across represented groups:
\begin{equation}
S_t=\frac{1}{|\mathcal{G}_t|}\sum_{g\in\mathcal{G}_t}
\frac{1}{|E_t\cap g|}\sum_{i\in E_t\cap g}s_i.
\end{equation}
 F1 averages positive-reference questions; empty-reference questions receive separate false-positive counts. Unavailable outcomes remain visible through the reported metric coverage.

T4 compares canonicalised prediction and reference MusicXML using NED, character error rate (CER), symbol error rate (SER), line error rate (LER) and normalised tree edit distance (TEDn). NED aggregates notation edits relative to the combined reference and prediction symbol counts. CER, SER and LER aggregate edits relative to their reference units in the derived ABC representation. TEDn averages normalised tree distances over items. Its all-parts variant removes trailing measures from each part until each input tree contains at most 6,000 nodes; prediction and reference can therefore retain different musical extents after truncation. Our NED comparison uses the canonicalised notation rather than requiring the unmodified MusicXML to satisfy the official OMR-NED input conditions.

The five distances use different units and normalisations, so their rankings need not agree. Parsing, conversion and computational limits can leave measurements unavailable; coverage remains explicit out of 125 questions. We assess the Opus--Astra ordering with 10,000 paired source-content-group bootstrap replicates. SM Tables~\ref{tab:nedsensitivity} and~\ref{tab:commonvalid} report alternative weightings and conditional analyses.

\subsection{Results}
Our analysis asks which operations distinguish models, whether transcription rankings persist across scopes, and how answer handling changes the interpretation of results. Table~\ref{tab:capabilities} presents a declared representative metric for every non-transcription subtype. Table~\ref{tab:transcription} summarises transcription NED; SM Table~\ref{tab:t4strict} reports all five metrics and their availability.

\begin{table*}[t]
\centering\footnotesize
\setlength{\tabcolsep}{1.0pt}
\caption{\textbf{Capability profiles across the five non-transcription aspects.} Strict group-macro scores in percent, higher is better; bold marks column maxima. T1 uses localisation F1. T2.1--2.2 use region F1, T2.3 exact answer and T2.4 relation F1. T3.1 and T3.4 use relation F1, T3.2 exact answer and T3.3 onset F1. T5.1--5.3 use exact answer, T5.4 traversal exactness and T6 joint answer/evidence success. T1.4/T3.1/T3.4 cover 113/96/103 positive-reference questions; empty-reference results appear in SM Table~\ref{tab:negativefp}. T6.2 reports retrieval (61 questions); counting (64) appears in SM Table~\ref{tab:t62counts}.}
\label{tab:capabilities}
\input{tables/capabilities_combined_strict}
\end{table*}

\input{tables/controls_main}

\subsubsection{Capabilities depend on the requested operation}
Visible-fact extraction is substantially stronger than pitch interpretation for several models. Opus reaches 99.0\% joint success on visible-fact questions in T6.1 but 56.0\% exact answer on written/sounding-pitch questions in T5.2. Astra reaches 100.0\% and 81.7\%, respectively. These profiles distinguish retrieving visible facts from selecting and interpreting notes.

Sol and Astra achieve 91.0\% and 89.8\% localisation F1 on T1.1, versus 7.9\% for Opus. Yet Opus reaches 94.0\% on measure regions and 76.1\% on onset relations, separating fine detection from larger-scale organisation. 

Repeat traversal adds a navigation requirement. Opus and Astra have similar T5.4 point estimates, at 79.5\% and 79.3\%; the paired interval leaves their ordering unresolved. The task tests expansion of a notated route, complementing sounding-event reconstruction in T5.3.

Structure and relation results expose additional differences (Table~\ref{tab:capabilities}). Qwen Max exceeds 84\% region F1 on systems/staves and measures, but reaches 46.9\% exact answer on part, staff and voice membership. Argon obtains 98.0\% continuation F1, compared with 57.7\% on tie/slur endpoints. The first contrast separates region geometry from object assignment; the second distinguishes continuity across page structures from the musical links between notes. Public targets let these questions examine association directly.

\begin{figure}[t]
\centering
\includegraphics[width=\linewidth]{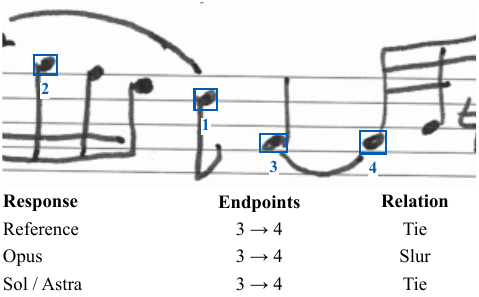}
\caption{\textbf{Correct endpoints, incorrect musical relation.} In this T3.1 example, Opus identifies endpoints 3 and 4 but returns a slur for the reference tie. Sol and Astra return the tie. Blue boxes reproduce all four public targets; the crop shows part of the original handwritten page presented to each model.}
\label{fig:relationcase}
\end{figure}

Figure~\ref{fig:relationcase} isolates relation typing: Opus identifies the correct endpoints but confuses a tie with a slur. A tie joins sounding duration, whereas a slur indicates phrasing.

\subsubsection{Model profiles vary across acquisition domains}
\begin{figure*}[t]
\centering
\includegraphics[width=\textwidth]{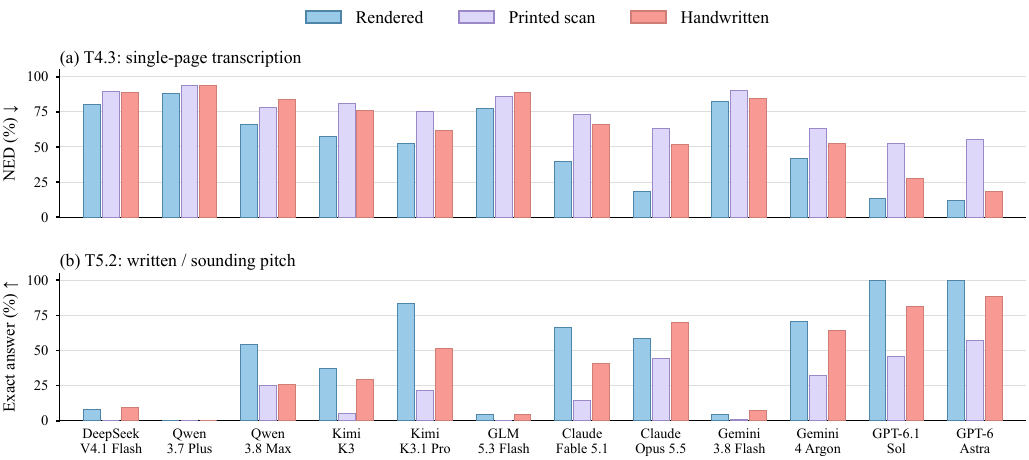}
\caption{\textbf{Model profiles vary across domains.} Each panel evaluates the same task across three domains for all twelve VLMs. T4.3 uses corpus NED (lower is better); T5.2 uses group-macro exact answer (higher is better). Model order stays fixed across panels.}
\label{fig:domains}
\end{figure*}

Figure~\ref{fig:domains} compares the same two tasks across Rendered, Printed scan and Handwritten inputs. On single-page transcription (T4.3), Astra has the lowest NED in Rendered and Handwritten, at 12.3\% and 18.2\%. Sol has the lowest Printed scan point estimate, at 52.3\%, followed by Astra at 55.3\%. The lowest NED thus occurs in different models across these evaluated domain subsets.

On written/sounding pitch (T5.2), Astra and Sol tie at 100.0\% on Rendered, while Astra leads Printed scan and Handwritten at 57.2\% and 88.7\%. Domain patterns also vary within a model: Opus scores higher on Handwritten than Rendered pitch questions, at 69.6\% versus 58.3\%, despite higher transcription error on Handwritten pages. These task-conditioned profiles distinguish visual notation reconstruction from pitch interpretation. The pitch point estimates rank Kimi K3.1 Pro third on Rendered but sixth on Printed scan; Opus moves from sixth to third. Domain-specific ranks expose differences concealed by the pooled score. Domains differ in content and givens; SM Sec.~\ref{sec:domain-results} reports question/group counts, descriptive ranks and confidence intervals.

\subsubsection{Counting requires localised evidence}
T6.2 changes the model ordering between retrieving one accidental and counting all noteheads in scope. Qwen Max leads retrieval at 95.1\%, but reaches 35.2\% count F1 and 3.3\% evidence F1. Astra reaches 79.5\% count F1 and 94.2\% evidence F1. SM Sec.~\ref{sec:qaanalysis} separates the retrieval and counting populations.

\subsubsection{Image removal tests visual dependence}
We remove all score images for Sol and DeepSeek Flash on the same 270-question subset: 30 questions from T2.3 and each T5/T6 subtype. Text, coordinates, givens, output schemas and generation settings remain fixed. Table~\ref{tab:imageremoval} gives the paired scores; SM Sec.~\ref{sec:imageremoval} details selection.

Sol's written/sounding-pitch score falls from 83.3\% to zero, and repeat traversal falls from 92.0\% to zero. Visible-fact success falls from 100.0\% to 13.6\%. Output remains valid on 268 of 270 image-removed questions, locating most of the decrease in answer content. Residual performance nevertheless remains substantial on structural membership and fragment transformation, at 53.3\% and 60.0\%. Public text, geometry and model priors thus support some answers without score pixels.

DeepSeek also loses visible-fact and retrieval success, but several musical scores are already near zero with images. Such floor effects limit the diagnostic value of small differences. SM Table~\ref{tab:imageremovalfull} reports component scores and paired group-bootstrap intervals. 

\subsubsection{Auxiliary givens isolate specific assistance}
The auxiliary-input controls retain score images while removing one specified assistance field. T5.2 deletes extra measure boxes but preserves marked-note targets; T6.3 deletes supplied bar durations while retaining the line assignments and rhythmic rules. Both conditions use identical references and generation settings (Table~\ref{tab:controlsmain}).

Sol retains 95.0\% exact accuracy on T5.2 and answers all thirteen T6.3 questions correctly after the specified field is removed. DeepSeek's T5.2 accuracy falls from 15.0\% to 5.0\%, while its T6.3 joint score remains unchanged. SM Sec.~\ref{sec:controls_followup} traces per-question answer and format transitions behind these aggregates.

\begin{table}[t]
\centering\footnotesize
\setlength{\tabcolsep}{2.5pt}
\caption{\textbf{Score-image removal has task-dependent effects.} Strict group-macro percentages on identical selected questions; higher is better. Image retains all inputs; Removed omits every image part. Metrics follow Table~\ref{tab:capabilities}. $^{*}$T6.2 shows retrieval; the 16 counting questions and all paired intervals appear in SM Table~\ref{tab:imageremovalfull}.}
\label{tab:imageremoval}
\input{tables/image_removal_main}
\end{table}

\subsubsection{Transcription rankings change with scope}
The leading transcription model changes on multipage inputs (Table~\ref{tab:transcription}). Astra has the lowest complete-population NED on T4.1, T4.2 and T4.3. On T4.4, Opus reaches 39.9\%, compared with 50.6\% for Astra, 55.2\% for Sol and 67.0\% for Kimi K3.1 Pro. 

The Opus--Astra reversal persists under equal-question and equal-group weighting. For multipage corpus NED, the paired group-bootstrap interval for Opus minus Astra is $[-18.9,-2.4]$ percentage points across 122 groups. Opus also has lower NED in all three reference-length strata (SM Table~\ref{tab:t4length}). Differences in source content prevent attributing this ordering to page count alone.

On the 80 questions where both Opus and Astra produce valid output, their NED values are 33.3\% and 34.6\%. The paired interval for Astra minus Opus is $[-3.0,5.7]$ points across 79 content groups, so this subset does not establish a reliable ordering. Opus leads on delivered multipage transcription, including output failures; the common-valid results place the two models closer in musical reconstruction quality.

\subsubsection{Page context with a fixed transcription target}
The matched-context comparison appends same-source pages while keeping the target excerpt, reference XML and inherited musical context fixed. It covers ten T4.1 and seven T4.2 targets, with a 32,768-token output cap in both conditions.

Opus's corpus NED increases by 6.3 points on T4.1 and 3.4 points on T4.2, whereas Astra's decreases by 0.6 and 8.6 points. Every paired interval includes zero (Table~\ref{tab:controlsmain}). The direction varies by model, with uncertainty spanning both benefit and harm. These controls measure how surrounding pages affect reconstruction of a fixed target.

T4.4 TEDn is incomplete for every model; SM Table~\ref{tab:t4strict} reports metric coverage alongside all five distances.

\begin{table}[t]
\centering\small
\setlength{\tabcolsep}{3pt}
\caption{\textbf{Transcription rankings depend on scope.} Corpus NED multiplied by 100, lower is better; bold marks the best complete-population value. Every model/subtype NED aggregate covers all 125 questions. Corpus ratios combine edit counts and symbol counts across the subtype. All five metrics appear in SM Table~\ref{tab:t4strict}.}
\label{tab:transcription}
\input{tables/transcription_ned_strict}
\end{table}

\subsubsection{Thin-symbol localisation is precision-sensitive}
A one-pixel tolerance affects thin-symbol localisation more than notehead/rest localisation. For Sol, T1.3 rises from 58.9\% to 72.3\%, while T1.1 rises from 91.0\% to 91.9\%. Predictions remain fixed and strict scores remain primary; SM Table~\ref{tab:pixel} reports all models.

The displacement changes IoU more for narrow stems than wide noteheads; tolerance does not correct class or attachment errors.

\subsubsection{Format compliance and content correctness}
Format compliance can dominate strict outcomes. Qwen Plus has 341 scored questions under the strict contract and 2,824 after permitted outer-fence removal; GLM Flash changes from 1,930 to 2,825 (Table~\ref{tab:formats}). The transformation exposes the enclosed content without correcting musical fields; subtype scores appear in SM Tables~\ref{tab:visualfence} and~\ref{tab:semanticfence}.

\begin{table}[t]
\centering\small
\setlength{\tabcolsep}{3pt}
\caption{\textbf{Removing a code fence changes scoreability.} Counts out of 3,000, not correct-answer counts. ``Scored'' denotes task-level scoreability; ``Failure'' denotes model-output failure. The remaining items have partially unresolved metrics. SM Tables~\ref{tab:t4strict} and~\ref{tab:t4fence} report native metric availability.}
\label{tab:formats}
\input{tables/format_outcomes}
\end{table}

\begin{figure}[t]
\centering
\includegraphics[width=\linewidth]{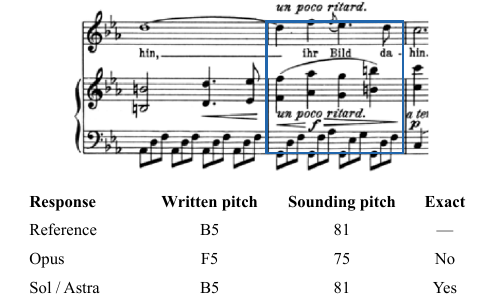}
\caption{\textbf{A valid format can contain a musical error.} T5.2 asks for the highest written pitch on the upper piano staff in the marked measure, with sounding pitch defined as written pitch minus two semitones (C4=60). The crop shows part of the fifth input page; the models receive all five pages. The blue outline reproduces the public measure scope.}
\label{fig:pitchcase}
\end{figure}

A concrete pitch question illustrates the distinction (Fig.~\ref{fig:pitchcase}). Opus returns valid JSON and applies the supplied transposition consistently to its chosen F5, but the reference target is B5. The format check succeeds while written and sounding pitch both fail. Sol and Astra return B5 and the correct sounding value. The example separates a consistent transposition calculation from selecting the correct note in the score.

\section{Conclusion}
\label{sec:conclusion}
\bench evaluates vision-language models through six aspects of musical score reading. Twelve VLMs exhibit distinct bottlenecks in fine localisation, musical interpretation and transcription. Image removal separates visual dependence from residual success without score pixels. Paired controls examine specific auxiliary fields and added page context while preserving the answer target. Format analysis separates output compatibility from musical content. These measurements support choosing VLMs for specific score-reading tasks and identify the visual and musical operations that require improvement.

\clearpage
\section*{LLM Usage Statement}
LLM provides auxiliary assistance with linguistic refinement, text restructuring, and the formatting of text and tables, as well as code for experiment orchestration, analysis and verification. The authors determine the research questions, methodology, experimental design, statistical analysis and interpretation, and retain responsibility for all content decisions and for reviewing and verifying AI-assisted material. Separately, the twelve multimodal models named in the experiments produce the benchmark answers under evaluation.

{\small
\bibliographystyle{ieeenat_fullname}
\bibliography{main}
}

%% file: sec/0_abstract.tex
\begin{abstract}
Vision-language models (VLMs) can answer questions about musical scores, but their ability to recognise notation, recover its structure and interpret its musical meaning requires separate evaluation. Existing optical music recognition (OMR) resources establish detection and transcription measurements; a transcription score alone leaves the strengths and bottlenecks of a VLM score reader unresolved. We introduce OmniScore-Bench, an evaluation of OMR capabilities in VLMs through 3,000 questions spanning six aspects and 24 task subtypes. Each question specifies a visible scope, reference annotations and an answer schema, enabling automatic evaluation from symbol localisation to musical consistency and evidence-linked question answering. Our assessment of twelve VLMs reveals operation-specific capability profiles and different transcription leaders across excerpt, single-page and multipage tasks. Controlled analyses examine the roles of score images, auxiliary givens and additional page context. On the tested subset, image removal exposes visual dependence in pitch and repeat interpretation while measuring how much performance remains with public context alone. Separate format and content analyses distinguish answer-format violations from musical errors. OmniScore-Bench supports diagnosing which score-reading operations VLMs perform reliably and which require improvement.
\end{abstract}

%% file: sec/1_intro.tex
\section{Introduction}
\label{sec:introduction}
A small mark can change how a musical score is read. A dot beside a note changes its duration; a line connecting two notes can join their sounding duration or indicate phrasing. Interpreting either requires locating the mark, identifying its endpoints or attachment, and applying the corresponding musical rule. Optical music recognition (OMR) therefore connects visual perception to a structured description of music~\cite{calvo2021understanding}. Vision-language models (VLMs) offer a common interface for examining these abilities through reconstruction and questions about the score at multiple levels of detail.

Established OMR resources measure important parts of this problem. DeepScoresV2 supports musical-object detection, MUSCIMA++ supplies handwritten symbols and relationships, and monophonic and pianoform datasets support transcription~\cite{tuggener2021deepscoresv2,hajic2017muscima,calvo2018primus,rios2023pianoform}. Sheet Music Benchmark (SMB) combines scanned notation with standardised transcription evaluation and the OMR Normalised Edit Distance (OMR-NED)~\cite{martinez2025smb}. MSU-Bench examines complete-score understanding and reports localisation errors and hallucinations with visual inputs~\cite{dai2026msu}. These findings motivate diagnosing individual score-reading operations: a score-level answer alone cannot distinguish failures in locating notation, associating musical objects or interpreting their musical consequences.

Our organising principle is to make each requested operation separately observable. A symbol question asks for classes and coordinates; a relation question asks which identified objects are connected; a musical question specifies the context needed to interpret pitch or time. Explicit targets and scopes support task-appropriate comparisons with reference annotations, while controlled changes to images or givens test what information supports the answer. Separate format and content checks distinguish representation failures from musical errors.

We introduce \bench\ to evaluate OMR capabilities in VLMs across symbols, structure, relations, transcription, musical consistency and score question answering (Fig.~\ref{fig:benchmark}). The benchmark contains 3,000 questions, with equal counts for its 24 subtypes, drawn from Rendered, Printed scan and Handwritten notation. Scoped references and answer schemas support automatic scoring, including five complementary transcription metrics. The resource exposes intermediate operations alongside reconstruction, enabling a capability profile with a separate measurement for each task.

Our evaluation of twelve VLMs reveals that proficiency depends on the operation and required representation. Astra leads excerpt and single-page transcription by complete-population notation edit distance, while Opus leads full multipage reconstruction. In a paired image-removal analysis, Sol's written/sounding-pitch score falls from 83.3\% to zero, while some structural and transformation tasks retain substantial performance without images. Additional controls remove selected auxiliary givens or append same-source pages while preserving the target output scope. These interventions distinguish visual dependence and specific contextual assistance from differences between task populations.

Our contributions are threefold: a six-aspect collection of OMR questions with scoped references and answer schemas for VLM evaluation; a systematic assessment of twelve VLMs that identifies operation-specific strengths and bottlenecks; and controlled analyses of visual input, auxiliary givens and page context, complemented by format and content diagnostics. Together, they support selecting and improving VLMs for particular score-reading operations.

We will publicly release the benchmark and associated code, including the evaluation scripts.

%% file: tables/omr_focus_matrix.tex
\begin{tabularx}{\linewidth}{@{}l*{6}{>{\centering\arraybackslash}X}@{}}
\toprule
\multirow[c]{2}{*}{Dataset / benchmark} & T1 & T2 & T3 & T4 & T5 & T6 \\
 & Localisation & Structure & Relations & Transcription & Consistency & Score QA \\
\midrule
DeepScoresV2~\cite{tuggener2021deepscoresv2} & \evalyes & \evalno & \evalno & \evalno & \evalno & \evalno \\
MUSCIMA++~\cite{hajic2017muscima} & \evalyes & \evalyes & \evalyes & \evalyes & \evalno & \evalno \\
PrIMuS~\cite{calvo2018primus} & \evalno & \evalno & \evalno & \evalyes & \evalno & \evalno \\
GrandStaff~\cite{rios2023pianoform} & \evalno & \evalno & \evalno & \evalyes & \evalno & \evalno \\
SMB~\cite{martinez2025smb} & \evalno & \evalyes & \evalno & \evalyes & \evalno & \evalno \\
Muse OMR Benchmark~\cite{musegroup2025benchmark} & \evalno & \evalyes & \evalyes & \evalyes & \evalno & \evalno \\
MSU-Bench~\cite{dai2026msu} & \evalno & \evalno & \evalno & \evalno & \evalno & \evalyes \\
\midrule
\textbf{OmniScore-Bench} & \evalyes & \evalyes & \evalyes & \evalyes & \evalyes & \evalyes \\
\bottomrule
\end{tabularx}

%% file: tables/capabilities_combined_strict.tex
\begin{tabular*}{\textwidth}{@{\extracolsep{\fill}}lrrrrrrrrrrrrrrrrrrrr@{}}
\toprule
 & \multicolumn{4}{c}{T1: Symbols} & \multicolumn{4}{c}{T2: Structure} & \multicolumn{4}{c}{T3: Relations} & \multicolumn{4}{c}{T5: Consistency} & \multicolumn{4}{c}{T6: Score QA} \\
\cmidrule(lr){2-5}\cmidrule(lr){6-9}\cmidrule(lr){10-13}\cmidrule(lr){14-17}\cmidrule(l){18-21}
Model & 1.1 & 1.2 & 1.3 & 1.4 & 2.1 & 2.2 & 2.3 & 2.4 & 3.1 & 3.2 & 3.3 & 3.4 & 5.1 & 5.2 & 5.3 & 5.4 & 6.1 & 6.2 & 6.3 & 6.4 \\
\midrule
DeepSeek V4.1 Flash & 0.4 & 1.7 & 0.3 & 0.5 & 36.0 & 41.4 & 54.1 & 32.5 & 27.0 & 26.6 & 27.5 & 18.4 & 28.4 & 5.6 & 0.0 & 24.5 & 93.1 & 59.0 & 20.7 & 47.1 \\
Qwen 3.7 Plus & 0.0 & 0.0 & 0.0 & 0.0 & 0.0 & 0.0 & 0.0 & 0.3 & 0.0 & 0.0 & 0.0 & 0.0 & 0.0 & 0.0 & 0.0 & 0.0 & 42.5 & 9.8 & 0.6 & 0.7 \\
Qwen 3.8 Max & 7.5 & 37.0 & 4.1 & 41.3 & 86.5 & 84.4 & 46.9 & 91.1 & 65.4 & 52.5 & 70.2 & 52.2 & 58.3 & 37.1 & 8.2 & 62.3 & 97.3 & \textbf{95.1} & 54.5 & 62.0 \\
Kimi K3 & 20.9 & 29.9 & 8.2 & 48.8 & 91.7 & 86.9 & 66.0 & 88.3 & 52.4 & 47.2 & 60.9 & 49.5 & 57.2 & 23.8 & 13.1 & 58.5 & 97.7 & 39.3 & 43.7 & 55.4 \\
Kimi K3.1 Pro & 36.4 & 55.0 & 21.5 & 60.1 & 94.0 & 91.5 & 70.3 & 89.5 & 57.3 & 66.7 & 70.6 & 61.4 & 64.8 & 53.2 & 21.3 & 64.0 & 98.3 & 57.4 & 58.1 & 66.5 \\
GLM 5.3 Flash & 0.1 & 3.7 & 0.2 & 0.9 & 44.5 & 31.4 & 3.3 & 48.1 & 30.2 & 12.7 & 42.3 & 14.6 & 20.5 & 2.8 & 1.1 & 30.6 & 97.2 & 62.3 & 23.5 & 38.2 \\
Claude Fable 5.1 & 4.1 & 26.6 & 3.3 & 17.7 & 88.3 & 86.6 & 63.5 & 89.2 & 53.2 & 61.0 & 68.7 & 64.7 & 62.4 & 41.7 & 13.1 & 64.6 & 97.5 & 73.8 & 61.1 & 69.0 \\
Claude Opus 5.5 & 7.9 & 24.0 & 4.7 & 24.8 & 86.7 & 94.0 & 88.1 & 77.0 & 66.2 & 76.0 & 76.1 & 57.3 & 84.6 & 56.0 & 46.3 & \textbf{79.5} & 99.0 & 62.3 & 81.0 & 89.3 \\
Gemini 3.8 Flash & 0.2 & 12.6 & 0.4 & 16.9 & 24.4 & 21.7 & 17.4 & 17.9 & 5.3 & 22.9 & 18.6 & 5.9 & 27.4 & 3.8 & 2.0 & 32.2 & 80.1 & 70.5 & 25.6 & 48.1 \\
Gemini 4 Argon & 75.1 & 92.0 & 43.3 & 73.4 & 98.8 & 97.4 & 89.3 & 98.0 & 57.7 & 66.7 & 78.4 & 65.7 & 76.3 & 55.3 & 32.2 & 67.8 & \textbf{100.0} & 93.4 & \textbf{91.3} & 79.0 \\
GPT-6.1 Sol & \textbf{91.0} & 91.7 & 58.9 & 85.4 & \textbf{98.9} & 98.2 & 91.0 & 99.7 & 86.8 & 87.9 & 90.2 & 89.7 & 83.8 & 75.7 & 66.2 & 78.2 & 99.4 & 93.4 & 86.5 & \textbf{91.2} \\
GPT-6 Astra & 89.8 & \textbf{94.8} & \textbf{76.2} & \textbf{86.1} & 98.6 & \textbf{98.4} & \textbf{95.1} & \textbf{99.8} & \textbf{88.7} & \textbf{90.4} & \textbf{93.9} & \textbf{93.3} & \textbf{89.8} & \textbf{81.7} & \textbf{77.2} & 79.3 & \textbf{100.0} & 90.2 & 88.6 & 89.1 \\
\bottomrule
\end{tabular*}

%% file: tables/controls_main.tex
\begin{table*}[t]
\centering
\caption{\textbf{Auxiliary-field removal and added page context test distinct dependencies.}
Scores are percentages; $\Delta=B-A$ is in percentage points with paired 95\% source-group bootstrap intervals.
Each question represents a distinct group, so $n$ also gives the number of groups.
$V_A/V_B$ counts format-valid answers in the two conditions.
Primary rows retain all model-output failures.
All conditions use a 32,768-token output cap.}
\label{tab:controlsmain}
\small
\setlength{\tabcolsep}{4pt}
\begin{tabular}{llcrrlc}
\toprule
Subtype & Model & $n$ & A & B & $\Delta$ [95\% CI] & $V_A/V_B$ \\
\midrule
\multicolumn{7}{l}{\textbf{Specific assistance: A = full; B = one field removed. Exact/joint success $\uparrow$.}} \\
T5.2 & DeepSeek V4.1 Flash & 20 & 15.00 & 5.00 & $-10.00\;[-30.00,\;10.00]$ & 13/11 \\
T5.2 & GPT-6.1 Sol & 20 & 95.00 & 95.00 & $0.00\;[-15.00,\;15.00]$ & 20/20 \\
T6.3 & DeepSeek V4.1 Flash & 13 & 15.38 & 15.38 & $0.00\;[0.00,\;0.00]$ & 11/6 \\
T6.3 & GPT-6.1 Sol & 13 & 100.00 & 100.00 & $0.00\;[0.00,\;0.00]$ & 13/13 \\
\midrule
\multicolumn{7}{l}{\textbf{Fixed-target context: A = target only; B = target plus source pages. Corpus NED $\downarrow$.}} \\
T4.1 & Claude Opus 5.5 & 10 & 34.69 & 40.99 & $+6.30\;[-8.92,\;29.07]$ & 9/9 \\
T4.1 & GPT-6 Astra & 10 & 11.56 & 10.92 & $-0.64\;[-9.49,\;6.77]$ & 10/10 \\
T4.2 & Claude Opus 5.5 & 7 & 49.27 & 52.71 & $+3.44\;[-10.97,\;24.05]$ & 6/5 \\
T4.2 & GPT-6 Astra & 7 & 43.83 & 35.19 & $-8.64\;[-41.95,\;27.74]$ & 7/7 \\
\bottomrule
\end{tabular}
\end{table*}

%% file: tables/image_removal_main.tex
\begin{tabular}{@{}lr rrrr@{}}
\toprule
& & \multicolumn{2}{c}{GPT-6.1 Sol} & \multicolumn{2}{c}{DeepSeek Flash} \\
\cmidrule(lr){3-4}\cmidrule(l){5-6}
Subtype & $n$ & Image & Removed & Image & Removed \\
\midrule
T2.3 & 30 & 90.0 & 53.3 & 66.7 & 30.0 \\
T5.1 & 30 & 76.7 & 3.3 & 13.3 & 3.3 \\
T5.2 & 30 & 83.3 & 0.0 & 3.3 & 6.7 \\
T5.3 & 30 & 59.6 & 0.0 & 0.0 & 0.0 \\
T5.4 & 30 & 92.0 & 0.0 & 42.0 & 10.0 \\
T6.1 & 30 & 100.0 & 13.6 & 90.2 & 10.6 \\
T6.2$^{*}$ & 14 & 92.9 & 21.4 & 64.3 & 14.3 \\
T6.3 & 30 & 80.8 & 23.1 & 26.9 & 11.5 \\
T6.4 & 30 & 84.0 & 60.0 & 38.0 & 28.0 \\
\bottomrule
\end{tabular}

%% file: tables/transcription_ned_strict.tex
\begin{tabular}{@{}lrrrr@{}}
\toprule
Model & T4.1 & T4.2 & T4.3 & T4.4 \\
\midrule
DeepSeek V4.1 Flash & 89.6 & 86.1 & 86.4 & 89.0 \\
Qwen 3.7 Plus & 90.8 & 89.9 & 91.8 & 96.1 \\
Qwen 3.8 Max & 81.1 & 81.1 & 76.1 & 76.2 \\
Kimi K3 & 71.7 & 69.3 & 70.9 & 74.5 \\
Kimi K3.1 Pro & 64.4 & 64.1 & 62.8 & 67.0 \\
GLM 5.3 Flash & 89.6 & 86.9 & 83.7 & 93.1 \\
Claude Fable 5.1 & 64.4 & 69.2 & 59.3 & 77.1 \\
Claude Opus 5.5 & 46.3 & 55.1 & 43.5 & \textbf{39.9} \\
Gemini 3.8 Flash & 84.4 & 79.8 & 85.7 & 98.0 \\
Gemini 4 Argon & 46.2 & 29.0 & 52.0 & 88.3 \\
GPT-6.1 Sol & 28.0 & 23.5 & 30.5 & 55.2 \\
GPT-6 Astra & \textbf{23.5} & \textbf{21.4} & \textbf{26.8} & 50.6 \\
\bottomrule
\end{tabular}

%% file: tables/format_outcomes.tex
\begin{tabular}{@{}lrrrr@{}}
\toprule
Model & \multicolumn{2}{c}{Strict} & \multicolumn{2}{c}{Outer fence only} \\
 & Scored & Failure & Scored & Failure \\
\midrule
DeepSeek V4.1 Flash & 2947 & 53 & 2947 & 53 \\
Qwen 3.7 Plus & 341 & 2658 & 2824 & 173 \\
Qwen 3.8 Max & 2953 & 47 & 2953 & 47 \\
Kimi K3 & 2962 & 37 & 2962 & 37 \\
Kimi K3.1 Pro & 2966 & 32 & 2966 & 32 \\
GLM 5.3 Flash & 1930 & 1056 & 2825 & 161 \\
Claude Fable 5.1 & 2881 & 119 & 2883 & 117 \\
Claude Opus 5.5 & 2946 & 54 & 2948 & 52 \\
Gemini 3.8 Flash & 1749 & 1251 & 2607 & 393 \\
Gemini 4 Argon & 2834 & 165 & 2849 & 149 \\
GPT-6.1 Sol & 2962 & 34 & 2962 & 34 \\
GPT-6 Astra & 2952 & 47 & 2952 & 47 \\
\bottomrule
\end{tabular}

%% file: supplementary.tex
\clearpage
\setcounter{page}{1}
\renewcommand{\thepage}{S\arabic{page}}
\setcounter{section}{0}
\setcounter{table}{0}
\setcounter{figure}{0}
\setcounter{equation}{0}
\renewcommand{\thesection}{S\arabic{section}}
\renewcommand{\thesubsection}{\thesection.\arabic{subsection}}
\renewcommand{\thetable}{S\arabic{table}}
\renewcommand{\thefigure}{S\arabic{figure}}
\renewcommand{\theequation}{S\arabic{equation}}
\renewcommand{\theHsection}{SM.\arabic{section}}
\renewcommand{\theHtable}{SM.\arabic{table}}
\renewcommand{\theHfigure}{SM.\arabic{figure}}
\renewcommand{\theHequation}{SM.\arabic{equation}}
\maketitlesupplementary
\phantomsection
\label{sm:start}
\raggedbottom

\section{Source and Domain Allocations}
Table~\ref{tab:sources} lists the source collections. Collection names distinguish source images from the annotation resources used to form the questions~\cite{fornes2012cvc,hajic2017muscima,gotham2022lieder,musegroup2025benchmark,weiss2021winterreise,weiss2024winterreise,zalkow2019measures,intuidoc2018accidentals}.

\begin{table}[t]
\centering\small
\caption{Questions by source collection. One rendered question includes synthetic degradation and remains in the rendered domain.}
\label{tab:sources}
\begin{tabularx}{\linewidth}{@{}lYr@{}}
\toprule
Domain & Collection & Questions \\
\midrule
Rendered & Muse native corpus & 952 \\
Rendered & Additional Muse pairs & 4 \\
Rendered & OpenScore Lieder & 44 \\
Handwritten & CVC-MUSCIMA source images & 680 \\
Handwritten & MUSCIMA++ annotations & 320 \\
Printed scan & Schubert Winterreise / Peters scans & 543 \\
Printed scan & AudioLabs measure annotations & 369 \\
Printed scan & Additional IMSLP pairs & 27 \\
Printed scan & Accidental-detection release & 61 \\
\midrule
Total & & 3000 \\
\bottomrule
\end{tabularx}
\end{table}

\paragraph{Image inventory.}
The main evaluation contains 1,905 distinct original images with 4,535 appearances in question inputs. Declared auxiliary region crops add 95 distinct images and 96 appearances: 24 in T6.1, 24 in T6.3 and 48 in T6.4. The complete input asset inventory therefore contains 2,000 distinct images.

\section{Subtype Definitions}
Table~\ref{tab:taskcontracts} specifies the target, public context and measurement for each of the 24 subtypes. The following conventions resolve distinctions that affect exact answers.

The event and assertion attributes in Figure~\ref{fig:benchmark} of the main paper describe the annotation vocabulary. T5 evaluates the requested timing, pitch, sounding-event and traversal answers; it does not ask every model to return a common pass/fail assertion vector.

\paragraph{Time and pitch.}
Time values use exact rationals in the units stated by the question. Written pitch retains step, alteration and octave, so enharmonic spelling remains distinct. Sounding pitch uses the specified chromatic coordinate and transformation.

\paragraph{Target identity and meter.}
Target identifiers are present and unique wherever the question requires them. Meter answers preserve the printed numerals rather than reducing them as an arithmetic fraction. Repeat questions admit an explicit unknown answer only for the stated ambiguity or unsupported-control conditions.

\begin{table*}[tp]
\centering\small
\caption{Subtype targets, question context and strict measurements. Each row defines a distinct requested operation; its measurement retains the corresponding task semantics.}
\label{tab:taskcontracts}
\begin{tabularx}{\textwidth}{@{}p{0.06\textwidth}p{0.25\textwidth}Yp{0.16\textwidth}@{}}
\toprule
Type & Target & Scope and public context & Main measurement \\
\midrule
T1.1 & Noteheads and rests & Classes and boxes in the declared image region & Localisation F1 \\
T1.2 & Clefs and accidentals & Class-aware symbol boxes; applicable labels are fixed & Localisation F1 \\
T1.3 & Stems, flags, beams and dots & Thin rhythmic marks within the visible scope & Localisation F1 \\
T1.4 & Dynamics, articulations and ornaments & Supported expression labels and their regions & Localisation F1 \\
T2.1 & Systems and staves & Page structure and normalised regions & Region F1 \\
T2.2 & Measures & Physical measure regions in the input extent & Region F1 \\
T2.3 & Part, staff and voice membership & Declared objects and any supplied structural givens & Exact answer \\
T2.4 & Structural continuation & Associations within the supplied systems or pages & Relation F1 \\
T3.1 & Tie and slur endpoints & Typed endpoints; declared chord equivalence for slurs & Relation F1 \\
T3.2 & Chord and beam groups & Group membership over the declared object set & Exact answer \\
T3.3 & Cross-voice timing & Onset relationships under the specified timing context & Onset F1 \\
T3.4 & Typed connections & Ties, slurs, beams and beam hooks within the declared scope & Relation F1 \\
T4.1 & Monophonic transcription & Bounded excerpt; MusicXML output & Five transcription metrics \\
T4.2 & Single-staff polyphony & Bounded polyphonic excerpt; MusicXML output & Five transcription metrics \\
T4.3 & Single-page transcription & Displayed page extent, with declared inherited context & Five transcription metrics \\
T4.4 & Multipage transcription & Ordered supplied pages; not necessarily a complete work & Five transcription metrics \\
T5.1 & Meter and timing & Target events, exact time units and pickup conventions & Exact answer \\
T5.2 & Written and sounding pitch & Target rules and explicit transposition conditions & Exact answer \\
T5.3 & Sounding events & Declared targets and valid tie-merging rules & Exact answer \\
T5.4 & Repeat traversal & Bounded visible route and explicit unknown conditions & Traversal exactness \\
T6.1 & Visible fact extraction & 61 meter, 31 title, 23 tempo-text and 10 part-label questions & Joint success \\
T6.2 & Local counting and retrieval & 64 notehead-count questions; 61 conditional accidental queries & Count/evidence F1; retrieval joint success \\
T6.3 & Conditional musical analysis & 107 onset, 15 chord-template and 3 key-signature questions & Joint success \\
T6.4 & Fragment transformation & 61 vocal-measure and 64 note-fragment comparisons & Joint success \\
\bottomrule
\end{tabularx}
\end{table*}

\section{Scoring Details and Comparison Stability}
\label{sec:scoring-details}

\paragraph{Inference settings.}
Temperature and top-$p$ use service defaults.

\paragraph{Public inputs and evidence.}
For each question, we use the same original images and page order across models, together with the same question, public givens, normalised region or target coordinates and output schema. Region and target coordinates are supplied separately rather than drawn on the original score images. Any auxiliary region images accompany the originals. The shared instruction requests only the specified JSON object or standalone MusicXML, without Markdown fences, scope truncation, continuation, repair, tools or external retrieval. Public fields can define target identities, context or transformation rules, but reference answers are separate.

Region-ID citation and box localisation are distinct evidence requirements. When the IDs are supplied, citation F1 measures whether the answer names the required regions. Joint success additionally requires the requested musical or textual answer to be correct. For example, a T6.3 key-signature question supplies region A and a major-or-natural-minor hypothesis set; the model returns signed fifths, both tonic/mode candidates, whether the signature determines actual tonality, and region A. The hypothesis set and region ID are givens, while the visible accidentals determine the fifths and tonic candidates.

\paragraph{Relation equivalence.}
T3.1 compares ties and slurs; T3.4 also includes beams and beam hooks, restricted to the types requested by each question. A relation records its type, directed endpoints and applicable beam level or hook direction. Relation order is immaterial, but duplicate multiplicity remains. For slurs only, reference-defined, disjoint groups of marked notes in the same logical chord map to a common endpoint representative. The same map applies to prediction and reference before matching. Ties, beams and hooks retain their original endpoints; the prediction cannot define new equivalence groups.

\paragraph{Traversal success.}
Let $V_i$ denote valid output, $R_i$ exact route agreement and $E_i$ exact expanded-event multiset agreement. For a definite-route question, traversal success is
\[
s_i=V_i R_i\bigl[(1-a_i)+a_i E_i\bigr],
\]
where $a_i=1$ only when expanded events are required. For the unknown-reference question, success instead requires a valid unknown response, accepted reason codes and a reason that cites the required affected and evidence regions. A concrete route does not satisfy that question. The primary score averages these per-question successes within and then across content groups.
\paragraph{Canonical notation and failure handling.}
Both valid MusicXML inputs undergo the same three-stage cleaning. First, identification entries \texttt{encoding}, \texttt{source}, \texttt{relation} and \texttt{miscellaneous} are removed. Instrument IDs follow part/instrument order; MIDI-device text and port attributes are cleared, and MIDI-channel, name, volume, pan and elevation fields are removed. MIDI program and unpitched values remain. Second, credits explicitly assigned to pages other than page~1 are removed. Third, redundant title, number and creator metadata are removed only when the header order and creator types satisfy the fallback checks: one part list follows all headers; at most one work and defaults element exists; creator types belong to composer, poet, lyricist, translator or arranger. With an explicit positive page height, nonempty first-page credit words positioned above its midpoint suppress title/number/creator fallbacks. Otherwise, a unique nonempty movement title or number suppresses only the corresponding work field. Notes, durations, divisions, voices, lyrics and musical markings remain unchanged. These rules fix the symbolic representation used for comparison.

For question $i$, let $e_i$ be notation edits and $r_i,p_i$ the reference and prediction symbol counts. Corpus NED is $\sum_i e_i/\sum_i(r_i+p_i)$. A known-invalid prediction uses $p_i=0$ and, when reference symbol counting succeeds, $e_i=r_i$. Its individual NED is therefore one. CER, SER and LER use edit counts divided by reference character, symbol and line counts after MusicXML-to-ABC conversion. Invalid predictions use empty text; an empty string still supplies one line under the line splitter, so its LER need not equal one. Unknown conversion or distance results remain unavailable. TEDn uses the all-parts \texttt{lmx} variant, with prediction and reference separately reduced to at most 6,000 XML nodes by removing the last measure from each part repeatedly. The independently truncated trees can retain different musical extents. The distance is edit cost divided by reference cost; the computation has a 300-second limit. TEDn remains unavailable for invalid inputs.

\paragraph{Geometric tolerance.}
The supplemental T1 criterion maximises IoU over translations of each predicted box by at most one original-image pixel per axis. It does not resize or clip boxes. The class/page constraints, one-to-one assignment, IoU threshold of 0.5, input validation and region-of-interest rules remain the same. This is a sensitivity analysis of the matching criterion, with the model answers held fixed.

For two rectangular boxes with equal widths $w$ and equal heights, with only a horizontal displacement $d<w$, their IoU is $(w-d)/(w+d)$. At an IoU threshold of 0.5, a match requires $d\leq w/3$. The same coordinate deviation can therefore reject a narrow stem box while retaining a wider notehead box. Class and attachment errors remain separate from coordinate tolerance.

\begin{table}[t]
\centering\small
\setlength{\tabcolsep}{3pt}
\caption{\textbf{Small coordinate deviations affect localisation scores.} Strict and one-pixel-tolerant group-macro F1, in percent. Predictions are identical across the two scoring conditions.}
\label{tab:pixel}
\input{tables/pixel_sensitivity}
\end{table}

\paragraph{Uniform MusicXML normalization.}
Every model uses the same MusicXML preprocessing, validation and metric computation. Preprocessing preserves musical content and checks that the normalised representation retains the element order, attributes and character content. External resource access and parser recovery are disabled; malformed XML and violations of size or nesting limits remain failures. This rule also applies to the separate outer-fence analysis.

\paragraph{Content grouping and stability.}
Groups follow source identity: MuseScore source IDs, OpenScore score IDs, CVC source-page IDs across writers and annotation collections, scan-piece IDs, and mother-page IDs for accidental crops. Linked OpenScore scans/renderings and linked AudioLabs/Winterreise material share their declared source groups. Groups are aggregation units based on source identifiers, rather than verified distinct compositions. Selection satisfies the domain/subtype quotas, then samples round-robin across groups in deterministic order without model scores or answer content. Table~\ref{tab:nedsensitivity} compares equal-question and equal-group averages of per-question NED with the primary corpus-ratio comparison. A paired bootstrap samples the same groups for Opus and Astra, with replacement, retaining all their questions; it recomputes each corpus ratio for 10,000 replicates (seed 20261112). The reported intervals are the 2.5th and 97.5th percentiles of paired differences, conditional on the observed grouping. They do not account for uncertainty across repeated generations or establish simultaneous rankings for all models.

For T5.4, the same paired resampling procedure recomputes each model's group-macro score and their difference, again using 10,000 replicates from the seeded random stream. The Opus--Astra transcription ordering reverses on T4.4 under all three weighting choices. In contrast, the T5.4 group-macro difference is only 0.24 percentage points, with a paired interval of $[-7.6,8.0]$ across 78 groups. The wide interval leaves the relative repeat-traversal performance unresolved.

\begin{table}[t]
\centering\footnotesize
\setlength{\tabcolsep}{3pt}
\caption{NED weighting sensitivity for Opus and Astra. Scores and interval endpoints are multiplied by 100. Intervals concern the primary corpus-NED difference, Opus minus Astra; negative values favour Opus.}
\label{tab:nedsensitivity}
\input{tables/ned_sensitivity}
\end{table}

\paragraph{Length strata and termination.}
The 125 multipage questions contain 2--84 original pages (median 4) and 404--54,443 reference symbols (median 4,021). We sort questions by the common reference-symbol count, break ties by question identifier, and form three near-equal strata. All predictions remain included. Table~\ref{tab:t4length} shows lower Opus corpus NED in every stratum. Length covaries with source material and content, so this comparison describes performance across the observed lengths without isolating a causal length effect.

\begin{table}[t]
\centering\footnotesize
\setlength{\tabcolsep}{3pt}
\caption{Multipage transcription by reference length. Symbol and original-page ranges describe each stratum. Corpus NED is multiplied by 100; lower is better.}
\label{tab:t4length}
\input{tables/t4_length_analysis}
\end{table}

Termination labels and scoring failures measure different properties (Table~\ref{tab:t4finish}). Opus has 16 explicit length terminations and Astra has seven; all these answers fail output validation. Normal termination does not guarantee a usable score: Astra has 31 output failures among its 118 normal stops, while Opus has none among 108. These counts complement the length-stratified distances by separating completion status from the validity and musical accuracy of the returned notation. All failed outputs remain in the full-population NED calculation.

\begin{table}[t]
\centering\footnotesize
\setlength{\tabcolsep}{4pt}
\caption{Termination labels for all 125 multipage answers. Parentheses count final model-output failures within each label. Unavailable termination metadata remain unknown; they are not inferred from scoring outcomes.}
\label{tab:t4finish}
\input{tables/t4_finish_outcomes}
\end{table}

\paragraph{Conditioning on valid output.}
Table~\ref{tab:conditionalned} separates full-population NED from format-valid-only NED. These latter scores use model-dependent subsets and cannot provide a fair cross-model ranking. Table~\ref{tab:commonvalid} instead restricts both models to identical questions with valid output and resamples their common content groups. The resulting T4.4 interval spans zero. This outcome-conditioned analysis describes musical discrepancies on common accepted answers; full-population scores additionally include output failures.

\begin{table*}[t]
\centering\footnotesize
\setlength{\tabcolsep}{3pt}
\caption{Format-conditioned transcription analysis. All and Valid denote corpus NED times 100 on all 125 questions and on the model's format-valid subset; $n_v$ is that subset's size. Valid-only columns have different populations across models and are not a replacement ranking.}
\label{tab:conditionalned}
\input{tables/format_conditional_ned}
\end{table*}

\begin{table}[t]
\centering\footnotesize
\setlength{\tabcolsep}{2.8pt}
\caption{Paired format-valid questions. $n/G$ gives question/content-group counts. NED and the 95\% interval for Astra minus Opus are multiplied by 100. Intervals use 10,000 paired group-bootstrap replicates.}
\label{tab:commonvalid}
\input{tables/common_valid_ned}
\end{table}

\section{Complete Transcription Results}
Tables~\ref{tab:t4strict} and~\ref{tab:t4fence} report all five transcription metrics for each subtype under strict scoring and the supplemental code-fence condition. Coverage accompanies unavailable aggregates, and no subset mean replaces a full-population value.

\paragraph{T4.1: monophonic excerpts.}
Agreement between local sequence metrics does not extend uniformly to line-level comparison. Astra has CER 27.4 and SER 30.9, compared with Opus at 45.4 and 47.4. The LER ordering reverses: Opus obtains 40.0 and Astra 49.5. Astra also has the lowest complete TEDn at 13.3. Character and symbol distances favour Astra, while the line-level distance favours Opus. The metric profile distinguishes local sequence fidelity from line-level reconstruction.

\paragraph{T4.2: single-staff polyphony.}
The sequence and tree measurements agree on the ordering of Astra and Sol. Their CER values are 29.5 and 30.5, SER values are 28.2 and 29.6, and TEDn values are 28.8 and 33.0, respectively. LER again selects a different model: Argon reaches 44.6, below Astra's 48.6. Only Kimi K3, Sol and Astra have complete TEDn coverage for this subtype. The tree-distance comparison therefore supports an ordering among these complete populations, without assigning an aggregate rank to the other models.

\paragraph{T4.3: single-page transcription.}
Page-level reconstruction also separates notation distance from the derived sequence distances. Opus has lower NED than Argon, at 43.5 versus 52.0, whereas Argon has lower CER, at 44.5 versus 47.8. Astra reaches CER 33.8 and SER 36.0; its LER of 54.3 is close to Opus at 54.5. Each of Sol's ABC-derived metrics covers 123 of the 125 questions, and no model has complete TEDn coverage. The different NED, character and line orderings expose distinct aspects of page reconstruction.

\paragraph{T4.4: multipage transcription.}
The multipage result extends beyond the NED reversal in the main paper. Opus has the lowest complete CER, SER and LER, at 64.3, 63.6 and 84.4. Qwen Max reaches 79.8, 83.2 and 89.5 on the same three measurements. Opus's advantage thus persists across several representations, although the substantial residual distances show that its output remains far from the references. Five models lack complete ABC-derived aggregates, and TEDn coverage ranges from 35 to 118 questions. These missing distances preclude a complete twelve-model ranking for those metrics.

\paragraph{Presentation sensitivity in transcription.}
Removing an outer code fence reduces Qwen Plus NED from 91.8 to 88.9 on T4.3 and from 96.1 to 91.6 on T4.4. The changes recover enclosed transcriptions but leave large content discrepancies. GLM Flash changes by less than one NED point in each subtype. The recovered transcriptions retain large notation distances, so formatting accounts for only a small part of their transcription error.

\begin{table*}[tp]
\centering\small
\caption{All five strict transcription metrics, multiplied by 100; lower is better. Each subtype contains 125 questions. A dash with $(k/125)$ gives the number of available items when a complete-population aggregate is unavailable. NED/CER/SER/LER use corpus ratios; TEDn$_{6k}$ uses an item mean of the capped variant in Section~\ref{sec:scoring-details}. CER, SER and LER can exceed 100. Bold marks minima among available complete-population values.}
\label{tab:t4strict}
\input{tables/transcription_strict}
\end{table*}

\begin{table*}[tp]
\centering\small
\caption{All five transcription metrics after removal of the permitted outermost code fence, with content unchanged. Values are multiplied by 100; lower is better. NED/CER/SER/LER use corpus ratios and TEDn$_{6k}$ uses an item mean of the capped variant in Section~\ref{sec:scoring-details}. Each subtype contains 125 questions; a dash with $(k/125)$ gives coverage for an unavailable complete-population aggregate. Bold marks minima among complete-population values.}
\label{tab:t4fence}
\input{tables/transcription_outer_fence_only}
\end{table*}

\section{Non-transcription Results by Subtype}
The following analyses complement the strict comparison in the main paper with operation-specific contrasts and component measurements. All scores in this section are percentages unless stated otherwise. Strict results remain the default; Tables~\ref{tab:visualfence} and~\ref{tab:semanticfence} report the outer-fence condition. T1.4, T3.1 and T3.4 have 113, 96 and 103 positive-reference questions, respectively; their remaining questions require empty predictions and receive the separate false-positive analysis in Table~\ref{tab:negativefp}. T6.2 has two output contracts, which we analyse separately.

\begin{table*}[tp]
\centering\small
\setlength{\tabcolsep}{3pt}
\caption{Visual and relational scores after outer-fence removal. Percent group macro, higher is better; bold marks column maxima. T1 uses localisation F1; T2.1--2.2 region F1; T2.3 exact answer; T2.4, T3.1 and T3.4 relation F1; T3.2 exact answer; T3.3 onset F1. T1.4, T3.1 and T3.4 F1 cover 113, 96 and 103 positive-reference questions; Table~\ref{tab:negativefp} covers the remaining empty-reference questions.}
\label{tab:visualfence}
\input{tables/capabilities_visual_outer_fence_only}
\end{table*}

\subsection{Symbol localisation}
\label{sec:symbolanalysis}
\paragraph{T1.1: noteheads and rests.}
Weak localisation persists after presentation errors are removed. Qwen Plus rises from 0.0 to 0.1 F1, and Gemini Flash rises from 0.2 to 5.2. These changes leave both models far below the stronger localisers. The supplemental condition therefore distinguishes an unreadable answer wrapper from the harder requirement to deliver correctly classified and positioned noteheads and rests.

\paragraph{T1.2: clefs and accidentals.}
The ordering among strong localisers depends on the symbol family. Argon reaches 92.0 F1 and Sol 91.7, whereas their T1.1 scores are 75.1 and 91.0. Astra reaches 94.8 on T1.2. Argon approaches Sol on clefs and accidentals while retaining a much larger gap on noteheads and rests, revealing a family-specific localisation profile.

\paragraph{T1.3: rhythmic marks.}
The tolerance analysis leaves substantial residual detection error in several models. Kimi K3.1 Pro rises from 21.5 to 27.1 F1 with one-pixel tolerance, while Opus rises from 4.7 to 6.4. Small coordinate deviations therefore explain only part of these low scores. The tolerance addresses small coordinate deviations; missed marks and class errors require different corrections.

\paragraph{T1.4: expression symbols.}
Kimi K3.1 Pro obtains 60.1 F1 on the 113 positive-reference questions, compared with Qwen Max at 41.3. The 12 questions with no target expression symbol expose a separate behaviour: the respective strict false-positive counts are 0.88 and 4.38 per question under group-macro averaging. The same ordering therefore holds for detecting supported symbols and avoiding unsupported additions in this pair. The empty-reference cases test whether models add expression marks absent from the requested scope.

\subsection{Structure}
\label{sec:structureanalysis}
\paragraph{T2.1: systems and staves.}
Sol, Argon and Astra cluster at 98.9, 98.8 and 98.6 region F1. The differences among these three are small in this collection, while Gemini Flash rises from 24.4 to 68.8 after outer-fence removal. Coarse page organisation is therefore nearly saturated for the leading models but remains sensitive to both presentation and content for a weaker one. This task captures the page-layout component of the capability profile.

\paragraph{T2.2: measure regions.}
System and measure localisation need not follow the same within-model pattern. Opus reaches 94.0 region F1 on measures and 86.7 on systems/staves (T2.1), whereas Qwen Max reaches 84.4 and 86.5, respectively. Both scores use region F1, so this contrast identifies a difference between recovering large staff systems and segmenting individual measures.

\paragraph{T2.3: structural membership.}
Membership assignment separates models that otherwise locate regions well. Qwen Max reaches 46.9 exact answer and Kimi K3.1 Pro 70.3, while both exceed 84 region F1 on T2.1 and T2.2. Astra reaches 95.1 exact answer. The membership task requires the declared part, staff and voice assignments to agree with the reference, so plausible region geometry alone cannot satisfy it. Public structural givens make this a test of the requested assignment within the stated context.

\paragraph{T2.4: continuation.}
Gemini Flash shows a large presentation effect on structural associations, rising from 17.9 to 83.2 relation F1 after outer-fence removal. Its supplemental T3.1 tie/slur result remains 5.3. Its continuation answers contain substantially stronger associations than its tie/slur answers. The continuation score concerns links across the supplied structures, whose identities and scope can constrain the problem differently from musical endpoint relations.

\subsection{Musical relations}
\label{sec:relationanalysis}
\paragraph{T3.1: tie and slur endpoints.}
Opus obtains 66.2 relation F1 on the 96 positive-reference questions, compared with Argon at 57.7, despite Argon's stronger page-region results. Both average 0.10 false positives per question under strict scoring on the 29 empty-reference questions. Similar restraint on absent links therefore coexists with different recovery of present links. Matching the typed endpoints supplies information beyond detecting a curve or identifying the surrounding staff.

\paragraph{T3.2: chord and beam groups.}
Astra reaches 90.4 exact answer, Sol 87.9 and Opus 76.0. Exact group membership requires the complete requested grouping to match, so partially correct groups do not receive the same credit as a fully correct answer. Kimi K3.1 Pro reaches 66.7 compared with Kimi K3 at 47.2, a wider separation than their T3.1 relation-F1 difference. The larger grouping gap identifies a specific gain in assembling notes into chords and beams.

\paragraph{T3.3: cross-voice timing.}
Outer-fence removal raises Qwen Plus onset F1 from 0.0 to 37.8 and GLM Flash from 42.3 to 52.5. Their recovered answers therefore contain some correct timing relations, but still fall below Kimi K3 at 60.9 and Fable at 68.7 under the same supplemental condition. The remaining gap concerns the returned onset relations, beyond whether the structured answer is accessible to the scorer.

\paragraph{T3.4: additional typed connections.}
Fable reaches 64.7 relation F1 on the 103 positive-reference questions, exceeding Opus at 57.3; the ordering reverses on T3.1, where Opus leads. Both produce zero strict false positives on the 22 empty-reference questions. The contrast therefore appears in recovering present typed connections rather than in an observed difference on these absent-target cases. A single relation score would obscure this change in model ordering.

\paragraph{Empty-reference questions.}
Table~\ref{tab:negativefp} reports false positives per question, averaged within source-content groups and then across groups. It complements the positive-reference F1 columns without assigning an arbitrary perfect F1 to an empty answer. A low false-positive count is useful only alongside output validity and positive-reference performance: a model that returns an invalid or empty answer can avoid emitting evaluable false positives while failing the task. Qwen Plus illustrates this distinction through zero strict false positives and near-zero strict F1 on the corresponding positive populations.

\begin{table*}[t]
\centering\small
\caption{Empty-reference errors. Each cell reports group-macro false positives per question (lower is better), with model-output failures in parentheses. These populations complement the positive-reference F1 results. Zero false positives on an invalid answer do not indicate correct abstention.}
\label{tab:negativefp}
\input{tables/negative_reference_fp}
\end{table*}

\begin{table*}[tp]
\centering\small
\caption{Musical consistency and evidence-linked QA after outer-fence removal. Percent group macro, higher is better; bold marks column maxima. T5.1--5.3 use exact answer, T5.4 traversal exactness and T6 joint success. T6.2 covers the 61 conditional-retrieval questions; Table~\ref{tab:t62counts} reports the other 64 counting questions.}
\label{tab:semanticfence}
\input{tables/capabilities_semantic_outer_fence_only}
\end{table*}

\subsection{Musical consistency}
\label{sec:consistencyanalysis}
\paragraph{T5.1: meter and timing.}
Component scores separate identifying meter from reconstructing event timing in T5.1. Qwen Max reaches 94.8\% meter exactness, but 74.2\% onset accuracy and 73.5\% duration accuracy; its complete-answer score is 58.3\%. Opus shows the same pattern at a higher level, with 97.6\% meter exactness, 90.1\% onset accuracy and 84.6\% complete-answer success. The meter is a shared context for the measure, while onset and duration must be recovered for each requested event. The component gaps locate the additional demands of temporal reconstruction.

\paragraph{T5.2: written and sounding pitch.}
The two pitch fields expose different error patterns within the same questions. Opus reaches 61.7 written-pitch accuracy and 56.2 sounding-pitch accuracy, with 56.0 complete-answer success. Sol's corresponding field scores are closer, at 76.5 and 76.2, with 75.7 complete-answer success. The field comparison separates two musical requirements: selecting a written pitch and returning its requested sounding coordinate impose related but distinct conditions. Field-level scores distinguish pitch selection from the requested sounding-pitch transformation.

\paragraph{T5.3: sounding events.}
T5.3 separates target coverage from sounding-event grouping. Astra, Sol and Opus all reach 100.0\% target-partition exactness, yet their event F1 values are 79.1\%, 67.9\% and 49.9\%. They include every requested target once while disagreeing on how notation joins those targets into sounding events. Event F1 remains close to each model's complete-answer success. Thus, event-level partial credit preserves the same broad performance pattern, concentrating the difficulty in reconstructing the musical groups rather than enumerating the targets.

\paragraph{T5.4: repeat traversal.}
Traversal components separate navigation from the expanded musical content. On the 124 questions with a definite route, Qwen Max reaches 91.7 measure-visit-sequence exactness. On the 23 questions that additionally require expanded written events, its event-multiset exactness is 0.0. Sol reaches 83.1 and 47.8 on these respective populations. The 23 questions form a rendered subset defined by reference visibility and supported event semantics, independently of model success. On their respective populations, the components distinguish route selection from reconstructing the expanded event sequence. The remaining question tests the explicit unknown condition rather than a definite route; none of the twelve models answers this case correctly.

\subsection{Evidence-linked score questions}
\label{sec:qaanalysis}
\paragraph{T6.1: visible fact extraction.}
The supplemental condition recovers visible-fact answers in models with strict formatting failures. Qwen Plus rises from 42.5 to 97.5 joint success, and Gemini Flash from 80.1 to 100.0. For Sol, strict region-citation F1 is 100.0 while answer accuracy and joint success are both 99.4. Thus, its remaining errors concern the answer under this contract rather than missing required region identifiers. The aggregate combines the fact types listed in Table~\ref{tab:taskcontracts}. The evidence component here checks selection of the supplied region identifiers.

\paragraph{T6.2: counting and conditional retrieval.}
This subtype contains two distinct populations. The 61 printed-scan questions ask for the accidental associated with a supplied notehead target and use joint answer/evidence success. The other 64 questions ask for filled and empty notehead counts with symbol-box evidence. The main table's joint-success column applies to the retrieval population only; Table~\ref{tab:t62counts} supplies the complementary count and evidence F1 values for all 64 counting questions.

The counting task also separates a count from its supporting locations on identical questions. Opus reaches 59.7\% count F1 but 10.3\% evidence F1, while Sol reaches 75.6\% and 85.8\%, respectively. Count F1 matches class/count facts; evidence F1 matches notehead boxes. The discrepancy for Opus shows that a plausible numerical answer can accompany weak localisation of its supporting symbols. This contrasts with T6.1 and T6.4, whose region-citation component selects supplied identifiers. Table~\ref{tab:t62counts} gives the complete comparison.

T6.2 contrasts retrieval for a given target with counting all relevant symbols. Qwen Max leads the 61 conditional accidental-retrieval questions at 95.1\% joint success, but reaches 35.2\% count F1 and 3.3\% evidence F1 on the 64 notehead-counting questions. Astra reaches 90.2\% retrieval success, 79.5\% count F1 and 94.2\% evidence F1. The model ordering changes with the requested operation: retrieving one accidental uses an identified target, whereas counting requires finding and distinguishing all filled and empty noteheads in scope.

 Table~\ref{tab:t62counts} extends the comparison to all twelve models and the outer-fence condition. Qwen Plus recovers 25.0 count F1 after fence removal while its evidence F1 remains zero; Gemini Flash recovers 16.5 and 0.7, respectively. Making the structured answer accessible therefore recovers some count facts while leaving symbol localisation weak.

\begin{table}[!t]
\centering\footnotesize
\setlength{\tabcolsep}{3pt}
\caption{Count and symbol-evidence F1 for all 64 T6.2 counting questions. Group-macro percentages, higher is better. The 61 conditional-retrieval questions are evaluated separately.}
\label{tab:t62counts}
\input{tables/t62_counting}
\end{table}

\paragraph{T6.3: conditional musical analysis.}
Argon leads this mixed analysis subtype at 91.3 joint success, ahead of Astra at 88.6 and Sol at 86.5. The order differs from T3.3 onset F1, where Astra and Sol exceed Argon. The subtype combines rhythmic comparisons with chord-template and key-signature questions. Its aggregate therefore measures success across these declared analyses, whereas T3.3 specifically measures onset reconstruction. The different populations and outputs support separate capability summaries.

\begin{table*}[t]
\centering\footnotesize
\setlength{\tabcolsep}{3pt}
\caption{Paired image-removal results for the fixed subset. Scores are strict group-macro percentages; $\Delta$ is removed-image minus image in percentage points. Intervals are pointwise 95\% paired content-group bootstrap intervals, without multiple-comparison correction. Metrics match Table~\ref{tab:capabilities} in the main paper except for the separately listed T6.2 count and evidence F1. Every listed measurement covers its complete selected population, including model-output failures.}
\label{tab:imageremovalfull}
\input{tables/image_removal_full}
\end{table*}

\paragraph{T6.4: fragment transformation.}
Sol reaches 91.2 joint success, compared with Opus at 89.3 and Astra at 89.1. All three have region-citation F1 of 100.0, and their answer-accuracy scores equal their joint scores. For these models, the remaining measured error therefore lies in the requested transformation answer, not failure to cite the required regions from A and B. Their similar point estimates accompany complete region citation, locating the remaining errors in transformation content.

\section{Image-Removal Comparison}
\label{sec:imageremoval}

\paragraph{Selection and intervention.}
We select 30 questions from each of T2.3, T5.1--5.4 and T6.1--6.4, for 270 questions shared by Sol and DeepSeek Flash. These structured-answer tasks test whether public targets, coordinates and interpretation rules support answers without score pixels. Selection uses acquisition domain and, for T6, question profile as strata. Each stratum contributes at least one question; the remaining allocation follows its remaining population, with deterministic source-group round-robin selection. Neither model scores nor answer content enter selection. The subset contains 88 Rendered, 98 Printed scan and 84 Handwritten questions. T6.2 comprises 14 retrieval and 16 counting questions.

The intervention removes every image part, including auxiliary crops, while preserving the question, output schema, target markers, region coordinates, public givens and textual image descriptions. It adds no instruction to guess or to compensate for missing images. The output cap remains 32,768 tokens. The image condition uses the main experiment's answers for the same question IDs; the removed-image condition requests one new answer per model and question. Output failures remain in the evaluation.

\paragraph{Paired analysis.}
Both conditions use the same strict scorer and references. For each subtype, we average scores within and then across the same source-content groups. Differences are removed-image minus image scores. Percentile intervals use 10,000 paired bootstrap samples of those groups, with all their questions retained. These intervals quantify content-sampling sensitivity for one answer per model and condition. T6.2 retrieval success, counting F1 and evidence F1 remain separate. The comparison measures dependence on score images in the presence of public text and geometry, rather than dependence on all visually derived information.

\paragraph{Component results and interpretation.}
Table~\ref{tab:imageremovalfull} extends the main comparison with counting and evidence localisation. For Sol, removing score images reduces T6.2 count F1 from 71.9 to 9.4 and symbol-evidence F1 from 87.2 to 0.3. The latter requires predicted boxes rather than selection of supplied region IDs. Public scope and coordinates therefore do not substitute for the missing notehead evidence on this subset. DeepSeek's corresponding image-condition scores are much lower, particularly its 0.6 evidence F1, which limits the absolute decrease that this intervention can expose.

The selected image-condition answers contain no format failures for Sol and two for DeepSeek; the removed-image answers contain two and 22, respectively. All these outcomes remain in the fixed populations. Sol's T5.2 and T5.4 answers are format-valid in both conditions, yet their removed-image scores are zero. These particular decreases concern answer content rather than representation validity. Conversely, DeepSeek's T5.3 score is zero in both conditions, and its T5.2 difference has an interval containing zero. These low-baseline results leave image dependence unresolved for those model/task pairs.

The selected questions are shared across conditions and models; group-macro scores apply to this fixed subset. Residual success without images can reflect public geometry, task constraints or model priors. The following controls isolate the contributions of two auxiliary fields while retaining the score images.

\section{Scores by Acquisition Domain}
\label{sec:domain-results}
Tables~\ref{tab:domain43} and~\ref{tab:domain52} give the scores behind Figure~\ref{fig:domains} in the main paper. Each domain uses the same task definition, scoring rule and question set for every model. T4.3 aggregates notation edit counts and symbol counts; T5.2 averages exact-answer scores within and then across source-content groups. We retain every selected question in its domain.

Ranks order point estimates separately within each task and domain, with shared ranks for exact ties. The intervals use 10,000 bootstrap resamples of source-content groups within each domain (seed 20261113), retaining each group's questions and recomputing the aggregate. They are pointwise 95\% percentile intervals for the observed content groups. The Printed scan T4.3 intervals for Sol and Astra overlap, so their point-estimate ordering is descriptive. Domain comparisons involve different source material and task givens, and therefore describe performance on the evaluated domain subsets rather than a controlled rendering change to identical scores.

\begin{table*}[t]
\centering\footnotesize
\setlength{\tabcolsep}{5pt}
\caption{T4.3 by domain: corpus NED in percent (lower is better), with 95\% intervals in brackets and domain-specific point-estimate ranks in parentheses. Bold marks the lowest point estimate.}
\label{tab:domain43}
\input{tables/domain_T43}
\end{table*}

\begin{table*}[t]
\centering\footnotesize
\setlength{\tabcolsep}{5pt}
\caption{T5.2 by domain: group-macro exact answer in percent (higher is better), with 95\% intervals in brackets and domain-specific point-estimate ranks in parentheses. Bold marks the highest point estimate, including ties.}
\label{tab:domain52}
\input{tables/domain_T52}
\end{table*}

\input{sec/controls_followup}

\input{sec/question_prompts}

\clearpage
\makeatletter
\global\@colroom\@colht
\global\vsize\@colht
\makeatother
{\small
\bibliographystyle{ieeenat_fullname}
\bibliography{main}
}

%% file: tables/pixel_sensitivity.tex
\begin{tabular}{@{}lrrrr@{}}
\toprule
Model & \multicolumn{2}{c}{T1.1} & \multicolumn{2}{c}{T1.3} \\
 & Strict & 1 px & Strict & 1 px \\
\midrule
DeepSeek V4.1 Flash & 0.4 & 0.6 & 0.3 & 0.4 \\
Qwen 3.7 Plus & 0.0 & 0.0 & 0.0 & 0.0 \\
Qwen 3.8 Max & 7.5 & 11.5 & 4.1 & 5.7 \\
Kimi K3 & 20.9 & 26.8 & 8.2 & 10.1 \\
Kimi K3.1 Pro & 36.4 & 45.2 & 21.5 & 27.1 \\
GLM 5.3 Flash & 0.1 & 0.1 & 0.2 & 0.4 \\
Claude Fable 5.1 & 4.1 & 5.8 & 3.3 & 4.8 \\
Claude Opus 5.5 & 7.9 & 10.6 & 4.7 & 6.4 \\
Gemini 3.8 Flash & 0.2 & 0.4 & 0.4 & 0.6 \\
Gemini 4 Argon & 75.1 & 80.1 & 43.3 & 67.2 \\
GPT-6.1 Sol & 91.0 & 91.9 & 58.9 & 72.3 \\
GPT-6 Astra & 89.8 & 90.7 & 76.2 & 88.3 \\
\bottomrule
\end{tabular}

%% file: tables/ned_sensitivity.tex
\begin{tabular}{@{}lrrrr@{}}
\toprule
 & & \multicolumn{2}{c}{Opus / Astra NED} & Difference \\
Type & Groups & Item mean & Group mean & 95\% interval \\
\midrule
T4.1 & 52 & 42.0 / 23.1 & 31.8 / 18.6 & [11.5, 28.4] \\
T4.2 & 49 & 49.4 / 25.5 & 40.0 / 18.1 & [18.6, 45.0] \\
T4.3 & 69 & 41.4 / 23.6 & 32.4 / 23.4 & [8.8, 23.6] \\
T4.4 & 122 & 33.1 / 50.1 & 33.4 / 49.7 & [-18.9, -2.4] \\
\bottomrule
\end{tabular}

%% file: tables/t4_length_analysis.tex
\begin{tabular}{@{}lrrrrr@{}}
\toprule
Length & $n$ & Ref. symbols & Pages & Opus & Astra \\
\midrule
Shorter & 42 & 404--2686 & 2--7 & 16.9 & 25.0 \\
Middle & 42 & 2724--4716 & 2--12 & 27.9 & 45.7 \\
Longer & 41 & 4720--54443 & 2--84 & 53.6 & 58.8 \\
\bottomrule
\end{tabular}

%% file: tables/t4_finish_outcomes.tex
\begin{tabular}{@{}lrrrr@{}}
\toprule
Model & Stop & Length & Unknown & Failures \\
\midrule
Opus & 108 (0) & 16 (16) & 1 (1) & 17 \\
Astra & 118 (31) & 7 (7) & 0 (0) & 38 \\
\bottomrule
\end{tabular}

%% file: tables/format_conditional_ned.tex
\begin{tabular}{@{}lrrrrrrrrrrrr@{}}
\toprule
Model & \multicolumn{3}{c}{T4.1} & \multicolumn{3}{c}{T4.2} & \multicolumn{3}{c}{T4.3} & \multicolumn{3}{c}{T4.4} \\ & All & $n_v$ & Valid & All & $n_v$ & Valid & All & $n_v$ & Valid & All & $n_v$ & Valid \\
\midrule
Claude Fable 5.1 & 64.4 & 124 & 64.0 & 69.2 & 121 & 68.3 & 59.3 & 122 & 58.7 & 77.1 & 60 & 53.1 \\
Claude Opus 5.5 & 46.3 & 125 & 46.3 & 55.1 & 123 & 54.2 & 43.5 & 124 & 43.0 & 39.9 & 108 & 31.0 \\
DeepSeek V4.1 Flash & 89.6 & 121 & 89.4 & 86.1 & 122 & 85.8 & 86.4 & 116 & 86.2 & 89.0 & 117 & 88.6 \\
Gemini 3.8 Flash & 84.4 & 106 & 83.0 & 79.8 & 110 & 78.3 & 85.7 & 80 & 77.1 & 98.0 & 35 & 94.5 \\
Gemini 4 Argon & 46.2 & 121 & 44.9 & 29.0 & 124 & 28.1 & 52.0 & 110 & 46.2 & 88.3 & 36 & 52.7 \\
Kimi K3 & 71.7 & 125 & 71.7 & 69.3 & 125 & 69.3 & 70.9 & 121 & 70.3 & 74.5 & 109 & 71.3 \\
Kimi K3.1 Pro & 64.4 & 125 & 64.4 & 64.1 & 124 & 63.6 & 62.8 & 124 & 62.5 & 67.0 & 111 & 62.4 \\
GPT-6 Astra & 23.5 & 125 & 23.5 & 21.4 & 125 & 21.4 & 26.8 & 119 & 22.8 & 50.6 & 88 & 35.2 \\
GPT-6.1 Sol & 28.0 & 124 & 27.4 & 23.5 & 125 & 23.5 & 30.5 & 121 & 27.9 & 55.2 & 96 & 42.4 \\
Qwen 3.7 Plus & 90.8 & 90 & 89.1 & 89.9 & 78 & 88.4 & 91.8 & 65 & 85.8 & 96.1 & 35 & 87.1 \\
Qwen 3.8 Max & 81.1 & 125 & 81.1 & 81.1 & 119 & 80.9 & 76.1 & 114 & 74.0 & 76.2 & 99 & 72.0 \\
GLM 5.3 Flash & 89.6 & 107 & 87.0 & 86.9 & 109 & 86.4 & 83.7 & 104 & 82.0 & 93.1 & 67 & 88.5 \\
\bottomrule
\end{tabular}

%% file: tables/common_valid_ned.tex
\begin{tabular}{@{}lrrrr@{}}
\toprule
Type & $n/G$ & Opus & Astra & $\Delta$ 95\% CI \\
\midrule
T4.1 & 125/52 & 46.3 & 23.5 & [-28.4, -12.1] \\
T4.2 & 123/49 & 54.2 & 22.0 & [-42.8, -18.6] \\
T4.3 & 118/66 & 42.0 & 22.0 & [-25.9, -13.3] \\
T4.4 & 80/79 & 33.3 & 34.6 & [-3.0, 5.7] \\
\bottomrule
\end{tabular}

%% file: tables/transcription_strict.tex
\begin{tabular}{@{}llrrrrr@{}}
\toprule
Task & Model & NED $\downarrow$ & CER $\downarrow$ & SER $\downarrow$ & LER $\downarrow$ & TEDn$_{6k}$ $\downarrow$ \\
\midrule
T4.1 & DeepSeek V4.1 Flash & 89.6 & 66.1 & 73.8 & 46.3 & \textemdash\,\scriptsize(121/125) \\
 & Qwen 3.7 Plus & 90.8 & 72.2 & 77.9 & 57.5 & \textemdash\,\scriptsize(91/125) \\
 & Qwen 3.8 Max & 81.1 & 59.6 & 65.8 & 43.7 & 42.6 \\
 & Kimi K3 & 71.7 & 57.8 & 61.9 & 45.3 & 43.0 \\
 & Kimi K3.1 Pro & 64.4 & 53.2 & 55.7 & 43.9 & 38.7 \\
 & GLM 5.3 Flash & 89.6 & 73.7 & 78.1 & 53.0 & \textemdash\,\scriptsize(107/125) \\
 & Claude Fable 5.1 & 64.4 & 51.9 & 55.8 & 42.0 & \textemdash\,\scriptsize(124/125) \\
 & Claude Opus 5.5 & 46.3 & 45.4 & 47.4 & \textbf{40.0} & 24.1 \\
 & Gemini 3.8 Flash & 84.4 & 58.9 & 64.3 & 45.0 & \textemdash\,\scriptsize(106/125) \\
 & Gemini 4 Argon & 46.2 & 34.3 & 37.8 & 40.5 & \textemdash\,\scriptsize(120/125) \\
 & GPT-6.1 Sol & 28.0 & 30.8 & 34.5 & 50.7 & \textemdash\,\scriptsize(124/125) \\
 & GPT-6 Astra & \textbf{23.5} & \textbf{27.4} & \textbf{30.9} & 49.5 & \textbf{13.3} \\
\midrule
T4.2 & DeepSeek V4.1 Flash & 86.1 & 67.6 & 69.6 & 58.9 & \textemdash\,\scriptsize(122/125) \\
 & Qwen 3.7 Plus & 89.9 & \textemdash\,\scriptsize(124/125) & \textemdash\,\scriptsize(124/125) & \textemdash\,\scriptsize(124/125) & \textemdash\,\scriptsize(93/125) \\
 & Qwen 3.8 Max & 81.1 & 59.9 & 62.0 & 54.4 & \textemdash\,\scriptsize(123/125) \\
 & Kimi K3 & 69.3 & 61.2 & 60.5 & 56.0 & 57.8 \\
 & Kimi K3.1 Pro & 64.1 & 59.0 & 57.2 & 55.1 & \textemdash\,\scriptsize(124/125) \\
 & GLM 5.3 Flash & 86.9 & \textemdash\,\scriptsize(124/125) & \textemdash\,\scriptsize(124/125) & \textemdash\,\scriptsize(124/125) & \textemdash\,\scriptsize(120/125) \\
 & Claude Fable 5.1 & 69.2 & 59.0 & 58.4 & 55.5 & \textemdash\,\scriptsize(121/125) \\
 & Claude Opus 5.5 & 55.1 & 51.9 & 50.4 & 49.3 & \textemdash\,\scriptsize(123/125) \\
 & Gemini 3.8 Flash & 79.8 & 61.7 & 64.2 & 55.9 & \textemdash\,\scriptsize(115/125) \\
 & Gemini 4 Argon & 29.0 & 34.0 & 32.8 & \textbf{44.6} & \textemdash\,\scriptsize(124/125) \\
 & GPT-6.1 Sol & 23.5 & 30.5 & 29.6 & 49.4 & 33.0 \\
 & GPT-6 Astra & \textbf{21.4} & \textbf{29.5} & \textbf{28.2} & 48.6 & \textbf{28.8} \\
\midrule
T4.3 & DeepSeek V4.1 Flash & 86.4 & 68.7 & 73.6 & 62.8 & \textemdash\,\scriptsize(118/125) \\
 & Qwen 3.7 Plus & 91.8 & 84.2 & 87.5 & 77.7 & \textemdash\,\scriptsize(66/125) \\
 & Qwen 3.8 Max & 76.1 & 65.8 & 70.3 & 62.8 & \textemdash\,\scriptsize(114/125) \\
 & Kimi K3 & 70.9 & 62.4 & 66.2 & 62.1 & \textemdash\,\scriptsize(121/125) \\
 & Kimi K3.1 Pro & 62.8 & 56.3 & 58.4 & 59.0 & \textemdash\,\scriptsize(124/125) \\
 & GLM 5.3 Flash & 83.7 & \textemdash\,\scriptsize(122/125) & \textemdash\,\scriptsize(122/125) & \textemdash\,\scriptsize(122/125) & \textemdash\,\scriptsize(103/125) \\
 & Claude Fable 5.1 & 59.3 & 52.5 & 54.5 & 56.4 & \textemdash\,\scriptsize(122/125) \\
 & Claude Opus 5.5 & 43.5 & 47.8 & 47.5 & 54.5 & \textemdash\,\scriptsize(124/125) \\
 & Gemini 3.8 Flash & 85.7 & 77.6 & 80.7 & 73.4 & \textemdash\,\scriptsize(80/125) \\
 & Gemini 4 Argon & 52.0 & 44.5 & 47.2 & 59.4 & \textemdash\,\scriptsize(110/125) \\
 & GPT-6.1 Sol & 30.5 & \textemdash\,\scriptsize(123/125) & \textemdash\,\scriptsize(123/125) & \textemdash\,\scriptsize(123/125) & \textemdash\,\scriptsize(121/125) \\
 & GPT-6 Astra & \textbf{26.8} & \textbf{33.8} & \textbf{36.0} & \textbf{54.3} & \textemdash\,\scriptsize(119/125) \\
\midrule
T4.4 & DeepSeek V4.1 Flash & 89.0 & 80.8 & 86.2 & 89.1 & \textemdash\,\scriptsize(118/125) \\
 & Qwen 3.7 Plus & 96.1 & 95.7 & 96.6 & 96.1 & \textemdash\,\scriptsize(36/125) \\
 & Qwen 3.8 Max & 76.2 & 79.8 & 83.2 & 89.5 & \textemdash\,\scriptsize(99/125) \\
 & Kimi K3 & 74.5 & \textemdash\,\scriptsize(124/125) & \textemdash\,\scriptsize(124/125) & \textemdash\,\scriptsize(124/125) & \textemdash\,\scriptsize(109/125) \\
 & Kimi K3.1 Pro & 67.0 & \textemdash\,\scriptsize(123/125) & \textemdash\,\scriptsize(123/125) & \textemdash\,\scriptsize(123/125) & \textemdash\,\scriptsize(111/125) \\
 & GLM 5.3 Flash & 93.1 & \textemdash\,\scriptsize(122/125) & \textemdash\,\scriptsize(122/125) & \textemdash\,\scriptsize(122/125) & \textemdash\,\scriptsize(65/125) \\
 & Claude Fable 5.1 & 77.1 & 86.7 & 87.4 & 91.6 & \textemdash\,\scriptsize(60/125) \\
 & Claude Opus 5.5 & \textbf{39.9} & \textbf{64.3} & \textbf{63.6} & \textbf{84.4} & \textemdash\,\scriptsize(108/125) \\
 & Gemini 3.8 Flash & 98.0 & 97.4 & 97.8 & 96.0 & \textemdash\,\scriptsize(35/125) \\
 & Gemini 4 Argon & 88.3 & 93.0 & 93.6 & 93.9 & \textemdash\,\scriptsize(36/125) \\
 & GPT-6.1 Sol & 55.2 & \textemdash\,\scriptsize(123/125) & \textemdash\,\scriptsize(123/125) & \textemdash\,\scriptsize(123/125) & \textemdash\,\scriptsize(96/125) \\
 & GPT-6 Astra & 50.6 & \textemdash\,\scriptsize(124/125) & \textemdash\,\scriptsize(124/125) & \textemdash\,\scriptsize(124/125) & \textemdash\,\scriptsize(88/125) \\
\bottomrule
\end{tabular}

%% file: tables/transcription_outer_fence_only.tex
\begin{tabular}{@{}llrrrrr@{}}
\toprule
Task & Model & NED $\downarrow$ & CER $\downarrow$ & SER $\downarrow$ & LER $\downarrow$ & TEDn$_{6k}$ $\downarrow$ \\
\midrule
T4.1 & DeepSeek V4.1 Flash & 89.6 & 66.1 & 73.8 & 46.3 & \textemdash\,\scriptsize(121/125) \\
 & Qwen 3.7 Plus & 90.1 & \textemdash\,\scriptsize(124/125) & \textemdash\,\scriptsize(124/125) & \textemdash\,\scriptsize(124/125) & \textemdash\,\scriptsize(116/125) \\
 & Qwen 3.8 Max & 81.1 & 59.6 & 65.8 & 43.7 & 42.6 \\
 & Kimi K3 & 71.7 & 57.8 & 61.9 & 45.3 & 43.0 \\
 & Kimi K3.1 Pro & 64.4 & 53.2 & 55.7 & 43.9 & 38.7 \\
 & GLM 5.3 Flash & 89.5 & 73.2 & 77.6 & 52.6 & \textemdash\,\scriptsize(108/125) \\
 & Claude Fable 5.1 & 64.4 & 51.9 & 55.8 & 42.0 & \textemdash\,\scriptsize(124/125) \\
 & Claude Opus 5.5 & 46.3 & 45.4 & 47.4 & \textbf{40.0} & 24.1 \\
 & Gemini 3.8 Flash & 84.4 & 58.5 & 63.9 & 44.6 & \textemdash\,\scriptsize(107/125) \\
 & Gemini 4 Argon & 46.3 & 34.2 & 37.7 & 40.2 & \textemdash\,\scriptsize(121/125) \\
 & GPT-6.1 Sol & 28.0 & 30.8 & 34.5 & 50.7 & \textemdash\,\scriptsize(124/125) \\
 & GPT-6 Astra & \textbf{23.5} & \textbf{27.4} & \textbf{30.9} & 49.5 & \textbf{13.3} \\
\midrule
T4.2 & DeepSeek V4.1 Flash & 86.1 & 67.6 & 69.6 & 58.9 & \textemdash\,\scriptsize(122/125) \\
 & Qwen 3.7 Plus & 88.3 & \textemdash\,\scriptsize(124/125) & \textemdash\,\scriptsize(124/125) & \textemdash\,\scriptsize(124/125) & \textemdash\,\scriptsize(116/125) \\
 & Qwen 3.8 Max & 81.1 & 59.9 & 62.0 & 54.4 & \textemdash\,\scriptsize(123/125) \\
 & Kimi K3 & 69.3 & 61.2 & 60.5 & 56.0 & 57.8 \\
 & Kimi K3.1 Pro & 64.1 & 59.0 & 57.2 & 55.1 & \textemdash\,\scriptsize(124/125) \\
 & GLM 5.3 Flash & 86.7 & \textemdash\,\scriptsize(124/125) & \textemdash\,\scriptsize(124/125) & \textemdash\,\scriptsize(124/125) & \textemdash\,\scriptsize(121/125) \\
 & Claude Fable 5.1 & 69.2 & 59.0 & 58.4 & 55.5 & \textemdash\,\scriptsize(121/125) \\
 & Claude Opus 5.5 & 55.1 & 51.9 & 50.4 & 49.3 & \textemdash\,\scriptsize(123/125) \\
 & Gemini 3.8 Flash & 79.8 & 61.5 & 64.1 & 55.5 & \textemdash\,\scriptsize(116/125) \\
 & Gemini 4 Argon & 29.0 & 34.0 & 32.8 & \textbf{44.6} & \textemdash\,\scriptsize(124/125) \\
 & GPT-6.1 Sol & 23.5 & 30.5 & 29.6 & 49.4 & 33.0 \\
 & GPT-6 Astra & \textbf{21.4} & \textbf{29.5} & \textbf{28.2} & 48.6 & \textbf{28.8} \\
\midrule
T4.3 & DeepSeek V4.1 Flash & 86.4 & 68.7 & 73.6 & 62.8 & \textemdash\,\scriptsize(118/125) \\
 & Qwen 3.7 Plus & 88.9 & 79.0 & 83.6 & 71.1 & \textemdash\,\scriptsize(90/125) \\
 & Qwen 3.8 Max & 76.1 & 65.8 & 70.3 & 62.8 & \textemdash\,\scriptsize(114/125) \\
 & Kimi K3 & 70.9 & 62.4 & 66.2 & 62.1 & \textemdash\,\scriptsize(121/125) \\
 & Kimi K3.1 Pro & 62.8 & 56.3 & 58.4 & 59.0 & \textemdash\,\scriptsize(124/125) \\
 & GLM 5.3 Flash & 83.3 & \textemdash\,\scriptsize(122/125) & \textemdash\,\scriptsize(122/125) & \textemdash\,\scriptsize(122/125) & \textemdash\,\scriptsize(105/125) \\
 & Claude Fable 5.1 & 59.3 & 52.5 & 54.5 & 56.4 & \textemdash\,\scriptsize(122/125) \\
 & Claude Opus 5.5 & 43.5 & 47.8 & 47.5 & 54.5 & \textemdash\,\scriptsize(124/125) \\
 & Gemini 3.8 Flash & 85.0 & 75.7 & 79.0 & 71.7 & \textemdash\,\scriptsize(86/125) \\
 & Gemini 4 Argon & 52.0 & \textemdash\,\scriptsize(124/125) & \textemdash\,\scriptsize(124/125) & \textemdash\,\scriptsize(124/125) & \textemdash\,\scriptsize(111/125) \\
 & GPT-6.1 Sol & 30.5 & \textemdash\,\scriptsize(123/125) & \textemdash\,\scriptsize(123/125) & \textemdash\,\scriptsize(123/125) & \textemdash\,\scriptsize(121/125) \\
 & GPT-6 Astra & \textbf{26.8} & \textbf{33.8} & \textbf{36.0} & \textbf{54.3} & \textemdash\,\scriptsize(119/125) \\
\midrule
T4.4 & DeepSeek V4.1 Flash & 89.0 & 80.8 & 86.2 & 89.1 & \textemdash\,\scriptsize(118/125) \\
 & Qwen 3.7 Plus & 91.6 & \textemdash\,\scriptsize(124/125) & \textemdash\,\scriptsize(124/125) & \textemdash\,\scriptsize(124/125) & \textemdash\,\scriptsize(91/125) \\
 & Qwen 3.8 Max & 76.2 & 79.8 & 83.2 & 89.5 & \textemdash\,\scriptsize(99/125) \\
 & Kimi K3 & 74.5 & \textemdash\,\scriptsize(124/125) & \textemdash\,\scriptsize(124/125) & \textemdash\,\scriptsize(124/125) & \textemdash\,\scriptsize(109/125) \\
 & Kimi K3.1 Pro & 67.0 & \textemdash\,\scriptsize(123/125) & \textemdash\,\scriptsize(123/125) & \textemdash\,\scriptsize(123/125) & \textemdash\,\scriptsize(111/125) \\
 & GLM 5.3 Flash & 92.9 & \textemdash\,\scriptsize(122/125) & \textemdash\,\scriptsize(122/125) & \textemdash\,\scriptsize(122/125) & \textemdash\,\scriptsize(67/125) \\
 & Claude Fable 5.1 & 77.1 & 86.7 & 87.4 & 91.6 & \textemdash\,\scriptsize(60/125) \\
 & Claude Opus 5.5 & \textbf{39.9} & \textbf{64.3} & \textbf{63.6} & \textbf{84.4} & \textemdash\,\scriptsize(108/125) \\
 & Gemini 3.8 Flash & 96.0 & 95.6 & 96.4 & 95.0 & \textemdash\,\scriptsize(45/125) \\
 & Gemini 4 Argon & 88.3 & 92.8 & 93.5 & 93.8 & \textemdash\,\scriptsize(36/125) \\
 & GPT-6.1 Sol & 55.2 & \textemdash\,\scriptsize(123/125) & \textemdash\,\scriptsize(123/125) & \textemdash\,\scriptsize(123/125) & \textemdash\,\scriptsize(96/125) \\
 & GPT-6 Astra & 50.6 & \textemdash\,\scriptsize(124/125) & \textemdash\,\scriptsize(124/125) & \textemdash\,\scriptsize(124/125) & \textemdash\,\scriptsize(88/125) \\
\bottomrule
\end{tabular}

%% file: tables/capabilities_visual_outer_fence_only.tex
\begin{tabular}{@{}lrrrrrrrrrrrr@{}}
\toprule
Model & T1.1 & T1.2 & T1.3 & T1.4 & T2.1 & T2.2 & T2.3 & T2.4 & T3.1 & T3.2 & T3.3 & T3.4 \\
\midrule
DeepSeek V4.1 Flash & 0.4 & 1.7 & 0.3 & 0.5 & 36.0 & 41.4 & 54.1 & 32.5 & 27.0 & 26.6 & 27.5 & 18.4 \\
Qwen 3.7 Plus & 0.1 & 0.6 & 0.8 & 2.2 & 30.7 & 35.1 & 12.7 & 37.5 & 21.7 & 14.7 & 37.8 & 10.6 \\
Qwen 3.8 Max & 7.5 & 37.0 & 4.1 & 41.3 & 86.5 & 84.4 & 46.9 & 91.1 & 65.4 & 52.5 & 70.2 & 52.2 \\
Kimi K3 & 20.9 & 29.9 & 8.2 & 48.8 & 91.7 & 86.9 & 66.0 & 88.3 & 52.4 & 47.2 & 60.9 & 49.5 \\
Kimi K3.1 Pro & 36.4 & 55.0 & 21.5 & 60.1 & 94.0 & 91.5 & 70.3 & 89.5 & 57.3 & 66.7 & 70.6 & 61.4 \\
GLM 5.3 Flash & 0.4 & 11.7 & 0.4 & 2.5 & 65.9 & 60.9 & 19.1 & 64.2 & 39.6 & 21.2 & 52.5 & 24.1 \\
Claude Fable 5.1 & 4.1 & 26.6 & 3.3 & 17.7 & 88.3 & 86.6 & 63.5 & 89.2 & 53.2 & 61.0 & 68.7 & 64.7 \\
Claude Opus 5.5 & 7.9 & 24.0 & 4.7 & 24.8 & 86.7 & 94.0 & 88.1 & 77.0 & 66.2 & 76.0 & 76.1 & 57.3 \\
Gemini 3.8 Flash & 5.2 & 24.4 & 5.3 & 29.7 & 68.8 & 67.2 & 24.8 & 83.2 & 5.3 & 24.0 & 24.7 & 10.6 \\
Gemini 4 Argon & 75.1 & 92.0 & 43.3 & 73.4 & 98.8 & 97.4 & 89.3 & 98.6 & 57.7 & 66.7 & 78.5 & 65.7 \\
GPT-6.1 Sol & \textbf{91.0} & 91.7 & 58.9 & 85.4 & \textbf{98.9} & 98.2 & 91.0 & 99.7 & 86.8 & 87.9 & 90.2 & 89.7 \\
GPT-6 Astra & 89.8 & \textbf{94.8} & \textbf{76.2} & \textbf{86.1} & 98.6 & \textbf{98.4} & \textbf{95.1} & \textbf{99.8} & \textbf{88.7} & \textbf{90.4} & \textbf{93.9} & \textbf{93.3} \\
\bottomrule
\end{tabular}

%% file: tables/negative_reference_fp.tex
\begin{tabular}{@{}lrrrrrr@{}}
\toprule
 & \multicolumn{2}{c}{T1.4 ($n=12$)} & \multicolumn{2}{c}{T3.1 ($n=29$)} & \multicolumn{2}{c}{T3.4 ($n=22$)} \\
\cmidrule(lr){2-3}\cmidrule(lr){4-5}\cmidrule(l){6-7}
Model & Strict & Outer fence only & Strict & Outer fence only & Strict & Outer fence only \\
\midrule
DeepSeek V4.1 Flash & 11.44 (0) & 11.44 (0) & 1.12 (0) & 1.12 (0) & 0.64 (0) & 0.64 (0) \\
Qwen 3.7 Plus & 0.00 (12) & 5.88 (0) & 0.00 (29) & 1.14 (2) & 0.00 (22) & 0.45 (3) \\
Qwen 3.8 Max & 4.38 (0) & 4.38 (0) & 0.39 (0) & 0.39 (0) & 0.05 (0) & 0.05 (0) \\
Kimi K3 & 1.56 (0) & 1.56 (0) & 0.54 (0) & 0.54 (0) & 0.14 (0) & 0.14 (0) \\
Kimi K3.1 Pro & 0.88 (0) & 0.88 (0) & 0.29 (0) & 0.29 (0) & 0.05 (0) & 0.05 (0) \\
GLM 5.3 Flash & 0.38 (2) & 1.94 (0) & 1.10 (6) & 1.28 (2) & 0.36 (6) & 0.68 (1) \\
Claude Fable 5.1 & 0.38 (0) & 0.38 (0) & 0.52 (0) & 0.52 (0) & 0.00 (0) & 0.00 (0) \\
Claude Opus 5.5 & 2.31 (1) & 4.44 (0) & 0.10 (0) & 0.10 (0) & 0.00 (0) & 0.00 (0) \\
Gemini 3.8 Flash & 0.81 (6) & 4.25 (1) & 0.32 (6) & 0.41 (0) & 0.09 (5) & 0.18 (0) \\
Gemini 4 Argon & 1.06 (0) & 1.06 (0) & 0.10 (0) & 0.10 (0) & 0.14 (0) & 0.14 (0) \\
GPT-6.1 Sol & 0.50 (0) & 0.50 (0) & 0.32 (0) & 0.32 (0) & 0.00 (0) & 0.00 (0) \\
GPT-6 Astra & 0.25 (0) & 0.25 (0) & 0.04 (0) & 0.04 (0) & 0.00 (0) & 0.00 (0) \\
\bottomrule
\end{tabular}

%% file: tables/capabilities_semantic_outer_fence_only.tex
\begin{tabular}{@{}lrrrrrrrr@{}}
\toprule
Model & T5.1 & T5.2 & T5.3 & T5.4 & T6.1 & T6.2 & T6.3 & T6.4 \\
\midrule
DeepSeek V4.1 Flash & 28.4 & 5.6 & 0.0 & 24.5 & 93.1 & 59.0 & 20.7 & 47.1 \\
Qwen 3.7 Plus & 26.6 & 11.1 & 0.7 & 18.3 & 97.5 & 45.9 & 22.3 & 36.7 \\
Qwen 3.8 Max & 58.3 & 37.1 & 8.2 & 62.3 & 97.3 & \textbf{95.1} & 54.5 & 62.0 \\
Kimi K3 & 57.2 & 23.8 & 13.1 & 58.5 & 97.7 & 39.3 & 43.7 & 55.4 \\
Kimi K3.1 Pro & 64.8 & 53.2 & 21.3 & 64.0 & 98.3 & 57.4 & 58.1 & 66.5 \\
GLM 5.3 Flash & 32.2 & 6.7 & 5.7 & 34.6 & 97.2 & 62.3 & 23.8 & 49.5 \\
Claude Fable 5.1 & 62.4 & 41.7 & 13.1 & 64.6 & 99.2 & 73.8 & 61.1 & 69.0 \\
Claude Opus 5.5 & 84.6 & 56.0 & 46.3 & \textbf{79.5} & 99.0 & 62.3 & 81.0 & 89.3 \\
Gemini 3.8 Flash & 32.6 & 4.7 & 2.0 & 41.4 & \textbf{100.0} & 80.3 & 33.9 & 53.2 \\
Gemini 4 Argon & 76.3 & 55.3 & 32.2 & 67.8 & \textbf{100.0} & 93.4 & \textbf{91.3} & 79.0 \\
GPT-6.1 Sol & 83.8 & 75.7 & 66.2 & 78.2 & 99.4 & 93.4 & 86.5 & \textbf{91.2} \\
GPT-6 Astra & \textbf{89.8} & \textbf{81.7} & \textbf{77.2} & 79.3 & \textbf{100.0} & 90.2 & 88.6 & 89.1 \\
\bottomrule
\end{tabular}

%% file: tables/t62_counting.tex
\begin{tabular}{@{}lrrrr@{}}
\toprule
 & \multicolumn{2}{c}{Strict} & \multicolumn{2}{c}{Outer fence only} \\
\cmidrule(lr){2-3}\cmidrule(l){4-5}
Model & Counts & Evidence & Counts & Evidence \\
\midrule
DeepSeek V4.1 Flash & 24.4 & 0.3 & 24.4 & 0.3 \\
Qwen 3.7 Plus & 0.0 & 0.0 & 25.0 & 0.0 \\
Qwen 3.8 Max & 35.2 & 3.3 & 35.2 & 3.3 \\
Kimi K3 & 39.8 & 21.4 & 39.8 & 21.4 \\
Kimi K3.1 Pro & 41.5 & 32.7 & 41.5 & 32.7 \\
GLM 5.3 Flash & 18.8 & 0.2 & 24.4 & 0.2 \\
Claude Fable 5.1 & 35.8 & 3.5 & 36.9 & 3.5 \\
Claude Opus 5.5 & 59.7 & 10.3 & 59.7 & 10.3 \\
Gemini 3.8 Flash & 4.0 & 0.0 & 16.5 & 0.7 \\
Gemini 4 Argon & 50.6 & 63.0 & 50.6 & 63.0 \\
GPT-6.1 Sol & 75.6 & 85.8 & 75.6 & 85.8 \\
GPT-6 Astra & 79.5 & 94.2 & 79.5 & 94.2 \\
\bottomrule
\end{tabular}

%% file: tables/image_removal_full.tex
\begin{tabular}{@{}lr rrrr rrrr@{}}
\toprule
& & \multicolumn{4}{c}{GPT-6.1 Sol} & \multicolumn{4}{c}{DeepSeek V4.1 Flash} \\
\cmidrule(lr){3-6}\cmidrule(l){7-10}
Subtype / measurement & $n$ & Image & Removed & $\Delta$ & 95\% CI & Image & Removed & $\Delta$ & 95\% CI \\
\midrule
T2.3 & 30 & 90.0 & 53.3 & -36.7 & $[-56.7, -16.7]$ & 66.7 & 30.0 & -36.7 & $[-56.7, -13.3]$ \\
T5.1 & 30 & 76.7 & 3.3 & -73.3 & $[-86.7, -56.7]$ & 13.3 & 3.3 & -10.0 & $[-23.3, 3.3]$ \\
T5.2 & 30 & 83.3 & 0.0 & -83.3 & $[-96.7, -70.0]$ & 3.3 & 6.7 & +3.3 & $[-6.7, 16.7]$ \\
T5.3 & 30 & 59.6 & 0.0 & -59.6 & $[-76.9, -40.4]$ & 0.0 & 0.0 & 0.0 & $[0.0, 0.0]$ \\
T5.4 & 30 & 92.0 & 0.0 & -92.0 & $[-100.0, -80.0]$ & 42.0 & 10.0 & -32.0 & $[-50.0, -16.0]$ \\
T6.1 & 30 & 100.0 & 13.6 & -86.4 & $[-100.0, -72.7]$ & 90.2 & 10.6 & -79.5 & $[-92.4, -64.4]$ \\
T6.2 retrieval & 14 & 92.9 & 21.4 & -71.4 & $[-92.9, -50.0]$ & 64.3 & 14.3 & -50.0 & $[-78.6, -21.4]$ \\
T6.2 count & 16 & 71.9 & 9.4 & -62.5 & $[-81.2, -40.6]$ & 21.9 & 18.8 & -3.1 & $[-18.8, 12.5]$ \\
T6.2 evidence & 16 & 87.2 & 0.3 & -86.9 & $[-93.2, -79.8]$ & 0.6 & 0.0 & -0.6 & $[-1.3, -0.1]$ \\
T6.3 & 30 & 80.8 & 23.1 & -57.7 & $[-78.8, -32.7]$ & 26.9 & 11.5 & -15.4 & $[-30.8, -3.8]$ \\
T6.4 & 30 & 84.0 & 60.0 & -24.0 & $[-44.0, -4.0]$ & 38.0 & 28.0 & -10.0 & $[-28.0, 6.0]$ \\
\bottomrule
\end{tabular}

%% file: tables/domain_T43.tex
\begin{tabular}{@{}lrrr@{}}
\toprule
Model & Rendered & Printed scan & Handwritten \\
\midrule
Questions / groups & 43 / 43 & 24 / 16 & 58 / 10 \\
\midrule
DeepSeek V4.1 Flash & 80.5 [76.5, 84.0] (10) & 89.9 [87.6, 91.8] (10) & 89.0 [86.5, 91.0] (11) \\
Qwen 3.7 Plus & 88.0 [83.6, 92.4] (12) & 93.9 [89.9, 97.7] (12) & 93.6 [91.5, 95.1] (12) \\
Qwen 3.8 Max & 66.2 [59.9, 72.1] (8) & 77.9 [72.9, 83.7] (7) & 84.0 [81.3, 86.6] (8) \\
Kimi K3 & 57.4 [51.6, 62.8] (7) & 80.7 [75.7, 84.9] (8) & 76.3 [71.0, 82.2] (7) \\
Kimi K3.1 Pro & 52.7 [46.9, 58.1] (6) & 75.6 [72.4, 79.5] (6) & 61.5 [54.7, 68.4] (5) \\
GLM 5.3 Flash & 77.6 [73.3, 82.0] (9) & 85.7 [82.1, 89.0] (9) & 88.7 [85.1, 91.3] (10) \\
Claude Fable 5.1 & 39.4 [33.4, 45.1] (4) & 73.0 [68.1, 78.7] (5) & 66.0 [59.8, 72.1] (6) \\
Claude Opus 5.5 & 18.1 [13.3, 23.9] (3) & 62.9 [55.2, 70.8] (3) & 51.9 [42.3, 61.2] (3) \\
Gemini 3.8 Flash & 82.4 [75.5, 88.8] (11) & 90.2 [83.2, 96.6] (11) & 84.8 [82.2, 88.1] (9) \\
Gemini 4 Argon & 41.9 [34.7, 49.6] (5) & 63.1 [55.5, 71.9] (4) & 52.3 [43.7, 61.1] (4) \\
GPT-6.1 Sol & 13.5 [5.5, 23.5] (2) & \textbf{52.3} [43.5, 60.4] (1) & 27.3 [21.1, 33.8] (2) \\
GPT-6 Astra & \textbf{12.3} [4.8, 21.6] (1) & 55.3 [41.5, 69.6] (2) & \textbf{18.2} [12.9, 24.4] (1) \\
\bottomrule
\end{tabular}

%% file: tables/domain_T52.tex
\begin{tabular}{@{}lrrr@{}}
\toprule
Model & Rendered & Printed scan & Handwritten \\
\midrule
Questions / groups & 24 / 24 & 61 / 22 & 40 / 14 \\
\midrule
DeepSeek V4.1 Flash & 8.3 [0.0, 20.8] (9) & 0.0 [0.0, 0.0] (10) & 9.5 [0.0, 21.4] (9) \\
Qwen 3.7 Plus & 0.0 [0.0, 0.0] (12) & 0.0 [0.0, 0.0] (10) & 0.0 [0.0, 0.0] (12) \\
Qwen 3.8 Max & 54.2 [33.3, 75.0] (7) & 25.4 [9.1, 43.6] (5) & 26.2 [11.9, 40.5] (8) \\
Kimi K3 & 37.5 [16.7, 58.3] (8) & 5.3 [0.0, 11.4] (8) & 29.2 [17.9, 40.5] (7) \\
Kimi K3.1 Pro & 83.3 [66.7, 95.8] (3) & 21.2 [7.6, 37.1] (6) & 51.8 [34.5, 69.0] (5) \\
GLM 5.3 Flash & 4.2 [0.0, 12.5] (10) & 0.0 [0.0, 0.0] (10) & 4.8 [0.0, 11.9] (11) \\
Claude Fable 5.1 & 66.7 [45.8, 83.3] (5) & 14.8 [4.2, 28.4] (7) & 41.1 [23.8, 59.5] (6) \\
Claude Opus 5.5 & 58.3 [37.5, 79.2] (6) & 44.7 [28.4, 61.4] (3) & 69.6 [50.0, 86.3] (3) \\
Gemini 3.8 Flash & 4.2 [0.0, 12.5] (10) & 1.1 [0.0, 3.4] (9) & 7.1 [0.0, 14.3] (10) \\
Gemini 4 Argon & 70.8 [50.0, 87.5] (4) & 32.6 [18.6, 48.1] (4) & 64.3 [50.0, 77.4] (4) \\
GPT-6.1 Sol & \textbf{100.0} [100.0, 100.0] (1) & 45.5 [30.7, 59.8] (2) & 81.5 [68.5, 93.5] (2) \\
GPT-6 Astra & \textbf{100.0} [100.0, 100.0] (1) & \textbf{57.2} [43.9, 69.7] (1) & \textbf{88.7} [76.8, 97.6] (1) \\
\bottomrule
\end{tabular}

%% file: sec/controls_followup.tex
\section{Matched Controls of Task Assistance and Page Context}
\label{sec:controls_followup}

We test whether specific task cues and surrounding score pages change answers when the target and reference remain fixed. The controls comprise 50 questions, each with two conditions and two models, for 200 answers. Both conditions use an output cap of 32,768 tokens; their results form a separate paired comparison.

\paragraph{Selection and scoring.}
Selection uses a fixed ordering independent of model outcomes and retains one question per source-content group. T5.2 uses the main exact-answer metric, and T6.3 uses joint answer/evidence success. These scores average within and across groups. Transcription uses corpus NED, computed from the summed native edit distances and symbol denominators. All model-output failures remain in the primary denominators. We use 10,000 paired source-group bootstrap samples to recompute each condition's score and its difference, with percentile 95\% intervals and no multiple-comparison correction. Each condition contains one answer per model and question, so the intervals describe variation across these selected groups rather than repeated inference on the same input.

\input{tables/controls_followup}

\paragraph{Extra measure boxes for marked notes.}
The T5.2 control removes only additional measure-region boxes. It preserves target-note boxes and IDs, original images, pitch conventions, the transposition rule and the output contract. Eligibility requires both explicit target markers and additional region boxes: 64 of the 125 questions meet this condition, spanning 38 groups. We select 20 groups, with ten Rendered and ten Handwritten questions. Profiles that use region-selection rules to define their targets fall outside this intervention.

\paragraph{Supplied aligned-measure duration.}
The T6.3 control deletes only \texttt{aligned\_measure\_duration}; it retains the two lines' target assignments, target boxes, measure regions, images and rhythmic rules. This field occurs in 46 questions across 13 groups. We select one question from every eligible group, yielding seven Rendered and six Handwritten questions. Profiles that instead use rule or template dictionaries, including all Printed scan questions, remain outside this comparison. Both controls estimate the contribution of the specified field on its eligible questions.

\paragraph{Additional pages with a fixed transcription target.}
The context control selects 17 distinct groups from 66 eligible T4.1/T4.2 questions. It compares the identical target crop alone with that crop followed by the full source-page sequence of the corresponding T4.4 task. Both conditions use the same score edition and identical target crop; the reference XML, inherited context and target-only output contract remain fixed. Both conditions explicitly restrict transcription to the original target crop. The ten T4.1 targets comprise four Rendered and six Printed scan questions; the seven T4.2 targets comprise two Rendered and five Printed scan questions. No Handwritten task meets this paired-page construction.

Table~\ref{tab:controls_followup} supplies the common-valid Opus diagnostics discussed in the main paper. These subsets condition on the format outcome in both conditions; they supplement rather than replace the full-population paired analysis.

\paragraph{Paired answer transitions.}
Removing an auxiliary field can preserve aggregate accuracy while changing which questions a VLM answers correctly. For Sol on T5.2, eighteen answers remain correct, one changes from correct to incorrect, and one changes in the opposite direction. Both conditions therefore score 95.0\%, although their successful question sets differ. All forty responses satisfy the output format, so these transitions concern answer content. On T6.3, Sol answers all thirteen questions correctly in both conditions. The images, target identities and remaining context support these observed scores after the specified field is removed.

DeepSeek shows a different response pattern. On T5.2, three previously correct answers become incorrect and one previously incorrect answer becomes correct, reducing accuracy from 15.0\% to 5.0\%. Four responses lose format validity and two gain it. On T6.3, the same two questions succeed in both conditions, while the other eleven remain unsuccessful. Nevertheless, six responses change from valid to invalid format and one changes from invalid to valid. The unchanged joint-success score therefore conceals changes in response compliance. Paired outcomes distinguish stable success, offsetting answer changes and persistent failures that aggregate scores alone can obscure.

\paragraph{Valid-output context comparison.}
Common-valid analysis identifies the role of usable outputs in this comparison. For Opus on T4.2, the five commonly valid questions improve from 48.4\% to 45.9\% NED, reversing the full-population direction. Astra's responses are valid throughout. Table~\ref{tab:controls_followup} gives the paired subset sizes and T4.1 results.

%% file: tables/controls_followup.tex
\begin{table}[t]
\centering\footnotesize
\caption{Common-valid context diagnostics for Opus. NED and differences are multiplied by 100; intervals use paired content-group bootstrap. These outcome-conditioned subsets supplement the full-population comparisons in the main paper.}
\label{tab:controls_followup}
\setlength{\tabcolsep}{3pt}
\begin{tabular}{@{}lrrrr@{}}
\toprule
Subtype & $n$ & Target & Context & $\Delta$ [95\% CI] \\
\midrule
T4.1 & 8 & 35.02 & 41.49 & $6.47\;[-2.57,25.35]$ \\
T4.2 & 5 & 48.43 & 45.87 & $-2.56\;[-14.42,11.78]$ \\
\bottomrule
\end{tabular}
\end{table}

%% file: sec/question_prompts.tex
\section{Question Prompts}

\label{sec:question-prompts}

The shared instruction below applies to every model. We then give one actual question text for each subtype, with both counting and retrieval examples for T6.2. Each question accompanies its instance-specific public givens, region coordinates, target identifiers, output schema and ordered score images. These fields specify the classes, target scope, musical assumptions and permitted answer structure; Table~\ref{tab:taskcontracts} summarises their roles. The public release includes the complete question instances.

\paragraph{Shared system instruction.}

\begin{quote}\small

Follow the supplied public task, givens and output contract exactly. Return only the requested JSON object, or raw standalone MusicXML text when the contract requests MusicXML. Do not add Markdown fences. Do not truncate the task scope to fit the output. No continuation, repair, tools or external retrieval is requested.

\end{quote}

\paragraph{User-message structure.}

The user message contains a JSON object with the fields \texttt{question}, \texttt{regions}, \texttt{target\_markers}, \texttt{givens} and \texttt{output\_contract}, followed by labelled original images and any declared auxiliary crops. The paragraphs below reproduce the \texttt{question} strings. Concrete coordinate arrays and JSON schemas accompany these strings in the model input; they are separate from the reference answers.

\subsection{Symbol localisation}

\paragraph{T1.1.}

\begin{quote}\small

Detect and classify every object from the supplied class list whose full bounding box lies within the supplied region. Return boxes in normalized original-page coordinates. Do not include objects crossing the region boundary.

\end{quote}

\paragraph{T1.2.}

\begin{quote}\small

Detect and classify every object from the supplied class list whose full bounding box lies within the supplied region. Return boxes in normalized original-page coordinates. Do not include objects crossing the region boundary.

\end{quote}

\paragraph{T1.3.}

\begin{quote}\small

Detect and classify every object from the supplied class list whose full bounding box lies inside the marked region. Return normalized original-page boxes. Use complete source glyphs, continuous beam strips and complete two-arm hairpins as specified; exclude objects crossing the region boundary.  Annotation profile: native engraving glyphs. A standalone flag is one complete flag glyph; its class (8th\_flag,16th\_flag,etc.) describes the complete glyph, and its box encloses the whole glyph. Do not split a multi-layer glyph into separate flag objects. Stem and beam drawing primitives are separate objects; secondary beam hooks are beams, not standalone flags.

\end{quote}

\paragraph{T1.4.}

\begin{quote}\small

Locate and classify all dynamic markings from the supplied class list wholly inside the supplied region. Use the declared complete-mark/text-block source profile, not character boxes. Return normalized original-page boxes; exclude objects crossing the region boundary.

\end{quote}

\subsection{Structure}

\paragraph{T2.1.}

\begin{quote}\small

Locate all music systems and five-line staves on this page. Return system/staff boxes and each staff's parent system. Staff boxes enclose outer line centres; do not include text or ledger lines.

\end{quote}

\paragraph{T2.2.}

\begin{quote}\small

Locate the displayed measure regions across each system. Return their boxes, parent system and left-to-right order. Keep a compressed multimeasure-rest display as one region with its indicated timed-measure span.

\end{quote}

\paragraph{T2.3.}

\begin{quote}\small

Group the marked notes by musical part, written staff and voice. Return memberships; group identifiers may be arbitrary but must be consistent. Distinguish simultaneous voices on the same staff.

\end{quote}

\paragraph{T2.4.}

\begin{quote}\small

Match each staff in the final system of the first page to its continuation in the first system of the next page. Return boxes for both boundary systems and every staff in those systems, a belongs\_to edge from each staff to its parent system, and a continues\_as edge from each staff on the first page to its matching staff on the next page. Use stable musical-part identity rather than vertical position alone. Staff boxes enclose the outer staff-line centres; system boxes enclose the staff envelope. Use the supplied output schema and your own unique local node IDs.

\end{quote}

\subsection{Musical relations}

\paragraph{T3.1.}

\begin{quote}\small

Among the marked events, return every direct endpoint relation of these types: slur, tie. For a tie or slur, use its true start/end notes; interior notes under the same curve are not extra endpoint pairs. A relation is directed from its notated start to its end. Give endpoints, type, direction and applicable beam level/hook attributes. Some items contain no requested relations: return an empty relations list in that case. Marker numbers carry no information about the answer.  Slur endpoint convention: when an endpoint is a notated chord, any marked notehead in that same chord and logical voice is an equivalent endpoint, including members of a cross-staff chord. Return one relation per slur, not additional copies for alternative chord members. When the geometrically closest head is unmarked, use another marked member of that endpoint chord; omit a relation if either endpoint has no marked member. This equivalence applies only to slurs. Ties remain specific to the connected written pitches; beam and beam-hook representative rules remain unchanged.

\end{quote}

\paragraph{T3.2.}

\begin{quote}\small

Group the marked note events by their written chord and by continuous beam group at each beam level. For chord\_groups, partition every marked note into groups that share a notated chord/stem; include singleton marked notes, and list marked rests separately as ungrouped\_rest\_targets. For beam\_groups, return each continuous group at each level that contains at least two marked note events. Report only the marked members; connections may pass through unmarked notes, and all marked heads of a beamed chord belong to that beam group. Level 1 is the primary eighth-note beam; level 2 is the secondary sixteenth-note beam, with higher levels continuing those duration subdivisions. Return beam hooks separately by level, forward/backward direction, and all marked heads sharing that hook stem. Group identifiers are unnecessary and member order is irrelevant. Do not return lowest-head representative edges or a chain of adjacent note pairs. An empty beam\_groups/hooks list is valid.

\end{quote}

\paragraph{T3.3.}

\begin{quote}\small

Read the marked notes in their common notated measure. Return each target and its onset from the measure start in exact quarter-note units. Notes belong to multiple voices; some or all may have different onsets. Ignore performance arpeggiation. Marker numbers carry no timing information.

\end{quote}

\paragraph{T3.4.}

\begin{quote}\small

Among the marked events, return every direct endpoint relation of these types: beam, beam\_hook, slur, tie. For a tie or slur, use its true start/end notes; interior notes under the same curve are not extra endpoint pairs. A relation is directed from its notated start to its end. For beam edges, use consecutive stem/chord groups within each continuous beam level, represented by the lowest written-pitch note in each complete chord; a beam hook is a self-relation at that representative, with level and forward/backward direction. These representative conventions apply only to this endpoint profile. Give endpoints, type, direction and applicable beam level/hook attributes. Some items contain no requested relations: return an empty relations list in that case. Marker numbers carry no information about the answer.  Slur endpoint convention: when an endpoint is a notated chord, any marked notehead in that same chord and logical voice is an equivalent endpoint, including members of a cross-staff chord. Return one relation per slur, not additional copies for alternative chord members. When the geometrically closest head is unmarked, use another marked member of that endpoint chord; omit a relation if either endpoint has no marked member. This equivalence applies only to slurs. Ties remain specific to the connected written pitches; beam and beam-hook representative rules remain unchanged.

\end{quote}

\subsection{Transcription}

\paragraph{T4.1.}

\begin{quote}\small

Transcribe this single-staff excerpt from an original printed scan into standalone score-partwise MusicXML. Use the supplied inherited musical context. Preserve exact note/rest onsets, durations, written pitches and voice membership. Include relations only when both endpoints lie inside this input scope; do not add off-scope measures or expand repeats. Output one part and staff 1; omit lyrics and free-position text/directions.

\end{quote}

\paragraph{T4.2.}

\begin{quote}\small

Transcribe this single-staff excerpt from an original printed scan into standalone score-partwise MusicXML. Use the supplied inherited musical context. Preserve exact note/rest onsets, durations, written pitches and voice membership. Include relations only when both endpoints lie inside this input scope; do not add off-scope measures or expand repeats. Output one part and staff 1; omit lyrics and free-position text/directions.

\end{quote}

\paragraph{T4.3.}

\begin{quote}\small

Transcribe all written notes and rests on this original scanned page into standalone score-partwise MusicXML. Use the supplied inherited musical context. Preserve exact note/rest onsets, durations, written pitches and voice membership. Include relations only when both endpoints lie inside this input scope; do not add off-scope measures or expand repeats. 

\end{quote}

\paragraph{T4.4.}

\begin{quote}\small

Transcribe all supplied pages, in their supplied order, as one score-partwise MusicXML score. Include the entire displayed source-file extent, all parts/staves, written events, relations and control notation. Do not infer missing movements or other material beyond these pages. Work-level completeness remains a review question.

\end{quote}

\subsection{Musical consistency}

\paragraph{T5.1.}

\begin{quote}\small

Read the marked handwritten events in the labelled notated measure. Return its active meter, nominal bar duration, and each marked event's onset and duration in exact quarter-note units. Respect dots and tuplets. Do not expand repeats or infer performed tempo.

\end{quote}

\paragraph{T5.2.}

\begin{quote}\small

For the marked notes in this handwritten page, report each written pitch as step, exact alteration and octave, and its sounding chromatic pitch. Read the visible key/accidental context. C4 is middle C and sounding C4=60. The supplied instrument condition is sounding pitch = written chromatic pitch + (0) semitones; this condition is given and is not inferred from unseen metadata.

\end{quote}

\paragraph{T5.3.}

\begin{quote}\small

Convert only the marked handwritten notes into sounding events for one written pass. Merge connected ties, retain separate attacks for untied notes, and ignore slurs, tempo and repeat playback. Return target-ID groups, onset, summed duration and sounding pitch; use quarter-note units from the beginning of the earliest labelled measure. The given sounding-minus-written transposition is 0 semitones, with C4=60.

\end{quote}

\paragraph{T5.4.}

\begin{quote}\small

Follow the visible notation within the marked measure regions. Start at the beginning of local-m1 with no prior traversal history. Apply any visible control at a boundary before deciding whether to continue or leave the marked range. Stop when normal forward progress leaves the end of the last marked measure, or a visible stopping instruction ends the traversal. Return the local measure IDs in visit order. Regions with the same local ID on different staff-line fragments form one notated measure. Unmarked material is context only: do not infer a missing or off-page jump destination; report unknown if the traversal requires one. The fixed safety limit is 256 measure visits. Return measure\_visit\_sequence under the attached JSON schema; expanded events are not requested. An unsupported or ambiguous traversal may use the schema's unknown response.

\end{quote}

\subsection{Score question answering}

\paragraph{T6.1.}

\begin{quote}\small

Read the visible time-signature notation in region A. For stacked numerals, return the upper and lower printed numbers as upper/lower. For the unbarred common-time letter sign, return C. Report what is visible: do not replace C with an inferred 4/4, reduce a printed fraction, or use catalogue/XML meter metadata. Cite region A.

\end{quote}

\paragraph{T6.2 (counting).}

\begin{quote}\small

Within the marked region, count filled and empty regular-size noteheads separately and return their boxes. Exclude grace/cue heads and objects crossing the region boundary.

\end{quote}

\paragraph{T6.2 (conditional retrieval).}

\begin{quote}\small

Find the explicit local accidental associated with the marked target notehead in this supplied contextual image. Return natural, sharp, flat, or none if no accidental is associated with that target. Other symbols or accidentals in the window are distractors. A positive answer must include one bounding box for the associated accidental; none must have an empty evidence list. Do not infer a key signature, a carried accidental from an earlier measure, or sounding pitch. This is an upstream preprocessed historical-score crop with staff lines apparently removed. The target notehead extends beyond the right crop boundary; the supplied target marker covers only its actually visible portion. Coordinates are normalized to this supplied crop, which is the complete available image.

\end{quote}

\paragraph{T6.3.}

\begin{quote}\small

In the complete written measure identified by A, compare note-attack onset sets on the unbraced vocal staff and the upper staff of the braced piano part. Collapse simultaneous chord notes to one attack; rests add no attack. Measure onsets from the start of this measure. Apply the supplied rule dictionary and cite A. The rectangle identifies the measure, while the original page supplies clef and grouping context. This profile contains no tied or grace targets.

\end{quote}

\paragraph{T6.4.}

\begin{quote}\small

Compare the vocal line on the top unbraced staff in the two complete written measures A and B. Include every written note and rest in temporal order; ignore lyrics, dynamics and articulations. Use chromatic pitch, not spelling, and measure onsets from each measure start. Apply the supplied transformation dictionary and cite both A and B. Each rectangle identifies a whole written measure on its original page; use surrounding clef/key/meter context. The profile has one vocal voice with no tied, grace or cue targets. A uniform time transformation multiplies every onset and duration by the same positive factor.

\end{quote}

%% file: main.bbl
\begin{thebibliography}{34}
\providecommand{\natexlab}[1]{#1}
\providecommand{\url}[1]{\texttt{#1}}
\expandafter\ifx\csname urlstyle\endcsname\relax
  \providecommand{\doi}[1]{doi: #1}\else
  \providecommand{\doi}{doi: \begingroup \urlstyle{rm}\Url}\fi

\bibitem[{Alibaba Cloud}(2026)]{qwen2026plus37}
{Alibaba Cloud}.
\newblock {Model Studio Model Release Notes: Qwen3.7-Plus}.
\newblock \url{https://www.alibabacloud.com/help/en/model-studio/newly-released-models}, 2026.
\newblock Official release entry dated 1 June 2026. Accessed 10 November 2026.

\bibitem[{Anthropic}(2026{\natexlab{a}})]{anthropic2026fable51}
{Anthropic}.
\newblock {Introducing Claude Fable 5.1 and Claude Mythos 5.1}.
\newblock \url{https://www.anthropic.com/claude-fable-and-mythos-5-1}, 2026{\natexlab{a}}.
\newblock Official release. Accessed 10 November 2026.

\bibitem[{Anthropic}(2026{\natexlab{b}})]{anthropic2026opus55}
{Anthropic}.
\newblock {Introducing Claude Opus 5.5}.
\newblock \url{https://www.anthropic.com/claude-opus-5-5}, 2026{\natexlab{b}}.
\newblock Official release, 22 September 2026. Accessed 10 November 2026.

\bibitem[Byrd and Simonsen(2015)]{byrd2015testbed}
Donald Byrd and Jakob~Grue Simonsen.
\newblock Towards a standard testbed for optical music recognition: Definitions, metrics, and page images.
\newblock \emph{Journal of New Music Research}, 44\penalty0 (3):\penalty0 169--195, 2015.

\bibitem[Calvo-Zaragoza and Rizo(2018)]{calvo2018primus}
Jorge Calvo-Zaragoza and David Rizo.
\newblock End-to-end neural optical music recognition of monophonic scores.
\newblock \emph{Applied Sciences}, 8\penalty0 (4):\penalty0 606, 2018.

\bibitem[Calvo-Zaragoza et~al.(2021)Calvo-Zaragoza, Haji{\v c}, and Pacha]{calvo2021understanding}
Jorge Calvo-Zaragoza, Jan Haji{\v c}, Jr., and Alexander Pacha.
\newblock Understanding optical music recognition.
\newblock \emph{ACM Computing Surveys}, 53\penalty0 (4):\penalty0 77:1--77:35, 2021.

\bibitem[Dai et~al.(2026)Dai, Yang, Li, Zhou, Liang, Zhang, Liu, Jin, An, Zhang, Jing, Lee, Zhang, Li, and Sun]{dai2026msu}
Congren Dai, Yue Yang, Krinos Li, Huichi Zhou, Shijie Liang, Bo Zhang, Enyang Liu, Ge Jin, Hongran An, Haosen Zhang, Peiyuan Jing, Kinhei Lee, Zhenxuan Zhang, Xiaobing Li, and Maosong Sun.
\newblock Musical score understanding benchmark: Evaluating large language models' comprehension of complete musical scores.
\newblock In \emph{Proceedings of the 64th Annual Meeting of the Association for Computational Linguistics (Volume 1: Long Papers)}, pages 10777--10799. Association for Computational Linguistics, 2026.

\bibitem[{DeepSeek-AI}(2026)]{deepseek2026flash1017}
{DeepSeek-AI}.
\newblock {DeepSeek-V4.1-Flash-1017}.
\newblock \url{https://huggingface.co/deepseek-ai/DeepSeek-V4.1-Flash-1017}, 2026.
\newblock Official model release. Accessed 10 November 2026.

\bibitem[Forn{\'e}s et~al.(2012)Forn{\'e}s, Dutta, Gordo, and Llad{\'o}s]{fornes2012cvc}
Alicia Forn{\'e}s, Anjan Dutta, Albert Gordo, and Josep Llad{\'o}s.
\newblock {CVC-MUSCIMA}: A ground truth of handwritten music score images for writer identification and staff removal.
\newblock \emph{International Journal on Document Analysis and Recognition}, 15\penalty0 (3):\penalty0 243--251, 2012.

\bibitem[Foscarin et~al.(2019)Foscarin, Jacquemard, and Fournier-S'niehotta]{foscarin2019diff}
Francesco Foscarin, Florent Jacquemard, and Rapha{\"e}l Fournier-S'niehotta.
\newblock A diff procedure for music score files.
\newblock In \emph{International Conference on Digital Libraries for Musicology}, pages 58--64, 2019.

\bibitem[Fu et~al.(2025)Fu, Kuang, Song, Huang, Yang, Li, Zhu, Luo, Wang, Lu, Li, Tang, Shan, Lin, Liu, Wu, Feng, Liu, Huang, Tang, Chen, Jin, Liu, and Bai]{fu2025ocrbenchv2}
Ling Fu, Zhebin Kuang, Jiajun Song, Mingxin Huang, Biao Yang, Yuzhe Li, Linghao Zhu, Qidi Luo, Xinyu Wang, Hao Lu, Zhang Li, Guozhi Tang, Bin Shan, Chunhui Lin, Qi Liu, Binghong Wu, Hao Feng, Hao Liu, Can Huang, Jingqun Tang, Wei Chen, Lianwen Jin, Yuliang Liu, and Xiang Bai.
\newblock {OCRBench v2}: An improved benchmark for evaluating large multimodal models on visual text localization and reasoning.
\newblock In \emph{Advances in Neural Information Processing Systems}, 2025.

\bibitem[{Google DeepMind}(2026{\natexlab{a}})]{google2026gemini38flash}
{Google DeepMind}.
\newblock {Gemini 3.8 Flash: Model Card}.
\newblock \url{https://deepmind.google/models/model-cards/gemini-3-8-flash/}, 2026{\natexlab{a}}.
\newblock Official model card, 2 September 2026. Accessed 10 November 2026.

\bibitem[{Google DeepMind}(2026{\natexlab{b}})]{google2026gemini4argon}
{Google DeepMind}.
\newblock {Gemini 4 Argon: Model Card}.
\newblock \url{https://deepmind.google/models/model-cards/gemini-4-argon/}, 2026{\natexlab{b}}.
\newblock Official model card, 13 October 2026. Accessed 10 November 2026.

\bibitem[Gotham and Jonas(2022)]{gotham2022lieder}
Mark Robert~Haigh Gotham and Peter Jonas.
\newblock The {OpenScore} lieder corpus.
\newblock In \emph{Music Encoding Conference Proceedings 2021}, pages 131--136. Humanities Commons, 2022.

\bibitem[Haji{\v c} and Pecina(2017)]{hajic2017muscima}
Jan Haji{\v c}, Jr. and Pavel Pecina.
\newblock The {MUSCIMA++} dataset for handwritten optical music recognition.
\newblock In \emph{International Conference on Document Analysis and Recognition}, pages 39--46, 2017.

\bibitem[{IntuiDoc}(2018)]{intuidoc2018accidentals}
{IntuiDoc}.
\newblock Accidental detection dataset, 2018.
\newblock Official dataset page, accessed 30 October 2026.

\bibitem[Martinez-Sevilla et~al.(2025)Martinez-Sevilla, Cerveto-Serrano, Luna, Chapman, Sapp, Rizo, and Calvo-Zaragoza]{martinez2025smb}
Juan~C. Martinez-Sevilla, Joan Cerveto-Serrano, Noelia Luna, Greg Chapman, Craig Sapp, David Rizo, and Jorge Calvo-Zaragoza.
\newblock Sheet music benchmark: Standardized optical music recognition evaluation, 2025.
\newblock Version 3, revised July 2026.

\bibitem[{Moonshot AI}(2026{\natexlab{a}})]{moonshot2026k3}
{Moonshot AI}.
\newblock {Kimi K3: Open Frontier Intelligence}.
\newblock \url{https://www.kimi.com/en/blog/kimi-k3}, 2026{\natexlab{a}}.
\newblock Official technical blog. Accessed 10 November 2026.

\bibitem[{Moonshot AI}(2026{\natexlab{b}})]{moonshot2026k31pro}
{Moonshot AI}.
\newblock {Kimi K3.1 Pro: Advancing Open Intelligence}.
\newblock \url{https://www.kimi.com/en/blog/kimi-k3-1-pro}, 2026{\natexlab{b}}.
\newblock Official technical blog. Accessed 10 November 2026.

\bibitem[{OpenAI}(2026{\natexlab{a}})]{openai2026astra}
{OpenAI}.
\newblock {GPT-6 Astra: A New Generation of Intelligence}.
\newblock \url{https://openai.com/index/gpt-6-astra/}, 2026{\natexlab{a}}.
\newblock Official release. Accessed 10 November 2026.

\bibitem[{OpenAI}(2026{\natexlab{b}})]{openai2026sol61}
{OpenAI}.
\newblock {Introducing GPT-6.1 Sol}.
\newblock \url{https://openai.com/index/introducing-gpt-6-1-sol/}, 2026{\natexlab{b}}.
\newblock Official release. Accessed 10 November 2026.

\bibitem[Ouyang et~al.(2025)Ouyang, Qu, Zhou, Zhu, Zhang, Lin, Wang, Zhao, Jiang, Zhao, Shi, Wu, Chu, Liu, Li, Xu, Zhang, Shi, Tu, and He]{ouyang2025omnidocbench}
Linke Ouyang, Yuan Qu, Hongbin Zhou, Jiawei Zhu, Rui Zhang, Qunshu Lin, Bin Wang, Zhiyuan Zhao, Man Jiang, Xiaomeng Zhao, Jin Shi, Fan Wu, Pei Chu, Minghao Liu, Zhenxiang Li, Chao Xu, Bo Zhang, Botian Shi, Zhongying Tu, and Conghui He.
\newblock {OmniDocBench}: Benchmarking diverse {PDF} document parsing with comprehensive annotations.
\newblock In \emph{Proceedings of the IEEE/CVF Conference on Computer Vision and Pattern Recognition}, pages 24838--24848, 2025.

\bibitem[Pereverzev and Abdullina(2025)]{musegroup2025benchmark}
Vasily Pereverzev and Kristina Abdullina.
\newblock {Muse OMR Benchmark}.
\newblock Dataset and evaluation software, 2025.
\newblock Dataset: musegroup/omr\_benchmark. Evaluator revision 8d06e7a.

\bibitem[Poznanski et~al.(2025{\natexlab{a}})Poznanski, Rangapur, Borchardt, Dunkelberger, Huff, Lin, Wilhelm, Lo, and Soldaini]{poznanski2025olmocr}
Jake Poznanski, Aman Rangapur, Jon Borchardt, Jason Dunkelberger, Regan Huff, Daniel Lin, Christopher Wilhelm, Kyle Lo, and Luca Soldaini.
\newblock {olmOCR}: Unlocking trillions of tokens in {PDFs} with vision language models.
\newblock \emph{arXiv preprint arXiv:2502.18443}, 2025{\natexlab{a}}.

\bibitem[Poznanski et~al.(2025{\natexlab{b}})Poznanski, Soldaini, and Lo]{poznanski2025olmocr2}
Jake Poznanski, Luca Soldaini, and Kyle Lo.
\newblock {olmOCR 2}: Unit test rewards for document {OCR}.
\newblock \emph{arXiv preprint arXiv:2510.19817}, 2025{\natexlab{b}}.

\bibitem[{Qwen Team}(2026)]{qwen2026max38}
{Qwen Team}.
\newblock {Qwen3.8-Max: A New Bar for Coding and Cowork}.
\newblock \url{https://qwen.ai/blog?id=qwen3.8}, 2026.
\newblock Official release, 3 August 2026. Accessed 10 November 2026.

\bibitem[R{\'i}os-Vila et~al.(2023)R{\'i}os-Vila, Rizo, I{\~n}esta, and Calvo-Zaragoza]{rios2023pianoform}
Antonio R{\'i}os-Vila, David Rizo, Jos{\'e}~M. I{\~n}esta, and Jorge Calvo-Zaragoza.
\newblock End-to-end optical music recognition for pianoform sheet music.
\newblock \emph{International Journal on Document Analysis and Recognition}, 26\penalty0 (3):\penalty0 347--362, 2023.

\bibitem[R{\'i}os-Vila et~al.(2026)R{\'i}os-Vila, Calvo-Zaragoza, Rizo, and Paquet]{rios2026fullpage}
Antonio R{\'i}os-Vila, Jorge Calvo-Zaragoza, David Rizo, and Thierry Paquet.
\newblock End-to-end full-page optical music recognition for pianoform sheet music.
\newblock \emph{International Journal of Computer Vision}, 134\penalty0 (2):\penalty0 49, 2026.

\bibitem[Tuggener et~al.(2018)Tuggener, Elezi, Schmidhuber, Pelillo, and Stadelmann]{tuggener2018deepscores}
Lukas Tuggener, Ismail Elezi, J{\"u}rgen Schmidhuber, Marcello Pelillo, and Thilo Stadelmann.
\newblock {DeepScores}: A dataset for segmentation, detection and classification of tiny objects.
\newblock In \emph{International Conference on Pattern Recognition}, pages 3704--3709, 2018.

\bibitem[Tuggener et~al.(2021)Tuggener, Satyawan, Pacha, Schmidhuber, and Stadelmann]{tuggener2021deepscoresv2}
Lukas Tuggener, Yvan~Putra Satyawan, Alexander Pacha, J{\"u}rgen Schmidhuber, and Thilo Stadelmann.
\newblock The {DeepScoresV2} dataset and benchmark for music object detection.
\newblock In \emph{International Conference on Pattern Recognition}, pages 9188--9195, 2021.

\bibitem[Wei{\ss} et~al.(2021)Wei{\ss}, Zalkow, Arifi-M{\"u}ller, M{\"u}ller, Koops, Volk, and Grohganz]{weiss2021winterreise}
Christof Wei{\ss}, Frank Zalkow, Vlora Arifi-M{\"u}ller, Meinard M{\"u}ller, Hendrik~Vincent Koops, Anja Volk, and Harald~G. Grohganz.
\newblock Schubert {Winterreise} dataset: A multimodal scenario for music analysis.
\newblock \emph{ACM Journal on Computing and Cultural Heritage}, 14\penalty0 (2):\penalty0 25:1--25:18, 2021.

\bibitem[Wei{\ss} et~al.(2024)Wei{\ss}, Zalkow, Arifi-M{\"u}ller, M{\"u}ller, Koops, Volk, and Grohganz]{weiss2024winterreise}
Christof Wei{\ss}, Frank Zalkow, Vlora Arifi-M{\"u}ller, Meinard M{\"u}ller, Hendrik~Vincent Koops, Anja Volk, and Harald~G. Grohganz.
\newblock Schubert {Winterreise} dataset, version 2.1.
\newblock \url{https://doi.org/10.5281/zenodo.10839767}, 2024.

\bibitem[{Z.ai}(2026)]{zai2026glm53flash}
{Z.ai}.
\newblock {GLM-5.3-Flash: Frontier Intelligence, Flash Cost}.
\newblock \url{https://z.ai/blog/glm-5.3-flash}, 2026.
\newblock Official release, 26 August 2026. Accessed 10 November 2026.

\bibitem[Zalkow et~al.(2019)Zalkow, Corrales, Tsai, Arifi-M{\"u}ller, and M{\"u}ller]{zalkow2019measures}
Frank Zalkow, Angel~Villar Corrales, TJ Tsai, Vlora Arifi-M{\"u}ller, and Meinard M{\"u}ller.
\newblock Tools for semi-automatic bounding box annotation of musical measures in sheet music.
\newblock In \emph{Demos and Late Breaking News of the International Society for Music Information Retrieval Conference (ISMIR)}, 2019.

\end{thebibliography}


\begin{thebibliography}{8}
\providecommand{\natexlab}[1]{#1}
\providecommand{\url}[1]{\texttt{#1}}
\expandafter\ifx\csname urlstyle\endcsname\relax
  \providecommand{\doi}[1]{doi: #1}\else
  \providecommand{\doi}{doi: \begingroup \urlstyle{rm}\Url}\fi

\bibitem[Forn{\'e}s et~al.(2012)Forn{\'e}s, Dutta, Gordo, and Llad{\'o}s]{fornes2012cvc}
Alicia Forn{\'e}s, Anjan Dutta, Albert Gordo, and Josep Llad{\'o}s.
\newblock {CVC-MUSCIMA}: A ground truth of handwritten music score images for writer identification and staff removal.
\newblock \emph{International Journal on Document Analysis and Recognition}, 15\penalty0 (3):\penalty0 243--251, 2012.

\bibitem[Gotham and Jonas(2022)]{gotham2022lieder}
Mark Robert~Haigh Gotham and Peter Jonas.
\newblock The {OpenScore} lieder corpus.
\newblock In \emph{Music Encoding Conference Proceedings 2021}, pages 131--136. Humanities Commons, 2022.

\bibitem[Haji{\v c} and Pecina(2017)]{hajic2017muscima}
Jan Haji{\v c}, Jr. and Pavel Pecina.
\newblock The {MUSCIMA++} dataset for handwritten optical music recognition.
\newblock In \emph{International Conference on Document Analysis and Recognition}, pages 39--46, 2017.

\bibitem[{IntuiDoc}(2018)]{intuidoc2018accidentals}
{IntuiDoc}.
\newblock Accidental detection dataset, 2018.
\newblock Official dataset page, accessed 30 October 2026.

\bibitem[Pereverzev and Abdullina(2025)]{musegroup2025benchmark}
Vasily Pereverzev and Kristina Abdullina.
\newblock {Muse OMR Benchmark}.
\newblock Dataset and evaluation software, 2025.
\newblock Dataset: musegroup/omr\_benchmark. Evaluator revision 8d06e7a.

\bibitem[Wei{\ss} et~al.(2021)Wei{\ss}, Zalkow, Arifi-M{\"u}ller, M{\"u}ller, Koops, Volk, and Grohganz]{weiss2021winterreise}
Christof Wei{\ss}, Frank Zalkow, Vlora Arifi-M{\"u}ller, Meinard M{\"u}ller, Hendrik~Vincent Koops, Anja Volk, and Harald~G. Grohganz.
\newblock Schubert {Winterreise} dataset: A multimodal scenario for music analysis.
\newblock \emph{ACM Journal on Computing and Cultural Heritage}, 14\penalty0 (2):\penalty0 25:1--25:18, 2021.

\bibitem[Wei{\ss} et~al.(2024)Wei{\ss}, Zalkow, Arifi-M{\"u}ller, M{\"u}ller, Koops, Volk, and Grohganz]{weiss2024winterreise}
Christof Wei{\ss}, Frank Zalkow, Vlora Arifi-M{\"u}ller, Meinard M{\"u}ller, Hendrik~Vincent Koops, Anja Volk, and Harald~G. Grohganz.
\newblock Schubert {Winterreise} dataset, version 2.1.
\newblock \url{https://doi.org/10.5281/zenodo.10839767}, 2024.

\bibitem[Zalkow et~al.(2019)Zalkow, Corrales, Tsai, Arifi-M{\"u}ller, and M{\"u}ller]{zalkow2019measures}
Frank Zalkow, Angel~Villar Corrales, TJ Tsai, Vlora Arifi-M{\"u}ller, and Meinard M{\"u}ller.
\newblock Tools for semi-automatic bounding box annotation of musical measures in sheet music.
\newblock In \emph{Demos and Late Breaking News of the International Society for Music Information Retrieval Conference (ISMIR)}, 2019.

\end{thebibliography}
